\documentclass{article}

\PassOptionsToPackage{numbers, sort&compress}{natbib}
\usepackage[preprint]{neurips_2024}

\usepackage[utf8]{inputenc} % allow utf-8 input
\usepackage[T1]{fontenc}    % use 8-bit T1 fonts
\usepackage{hyperref}       % hyperlinks
\usepackage{url}            % simple URL typesetting
\usepackage{booktabs}       % professional-quality tables
\usepackage{amsfonts}       % blackboard math symbols
\usepackage{nicefrac}       % compact symbols for 1/2, etc.
\usepackage{microtype}      % microtypography
\usepackage{xcolor}         % colors
\usepackage{multirow}
\usepackage{graphicx}
\usepackage{colortbl}
\usepackage[normalem]{ulem}
\useunder{\uline}{\ul}{}
\usepackage{kotex}
\usepackage{pdfpages}
\usepackage{float}

\usepackage[breakable,skins]{tcolorbox}
\arrayrulecolor{black}

\title{V{\Large ARCO}-V{\Large ISION}: Expanding Frontiers in \\ Korean Vision-Language Models}

\author{
    {Jeongho Ju\thanks{These authors contributed equally to this work. Authors are listed in alphabetical order.}
    \ \ \ \ \ \
    Daeyoung Kim\footnotemark[1]
    \ \ \ \ \ \
    SunYoung Park\footnotemark[1]
    \ \ \ \ \ \
    Youngjune Kim\footnotemark[1]\hspace{0.4em}\thanks{Corresponding author}} \\
  NC Research, NCSOFT \\
  \texttt{\{jeongho, daeyoungk, sun0park, youngjune\}@ncsoft.com}
}

\begin{document}

\maketitle

\newcommand\varcovision{{\normalsize V\small ARCO\normalsize-\normalsize V\small ISION}}
\newcommand\varcovisionfull{{\normalsize V\small ARCO\normalsize-\normalsize V\small ISION\normalsize -14B}}

\begin{abstract}
In this paper, we introduce an open-source Korean-English vision-language model (VLM), \varcovision. We incorporate a step-by-step training strategy that allows a model learn both linguistic and visual information while preserving the backbone model's knowledge. Our model demonstrates outstanding performance in diverse settings requiring bilingual image-text understanding and generation abilities compared to models of similar size. \varcovision~is also capable of grounding, referring, and OCR, expanding its usage and potential applications for real-world scenarios. In addition to the model, we release five Korean evaluation datasets, including four closed-set and one open-set benchmarks. We anticipate that our milestone will broaden the opportunities for AI researchers aiming to train VLMs. \varcovision~is available at \url{https://huggingface.co/NCSOFT/VARCO-VISION-14B}.

\end{abstract}

\section{Introduction}

Recent advancements in Large Language Models (LLMs) have increased attention on handling multimodality, providing robust backbones for Vision-Language Models (VLMs). Incorporating high-performing LLMs in VLMs demonstrated significant improvements across various visual tasks requiring text understanding, reasoning, and generation abilities~\citep{openai2024gpt4,team2023gemini,claude3,lu2024deepseek, abdin2024phi, li2024llava, laurenccon2024building, deitke2024molmo, wang2024qwen2, chen2024far, yao2024minicpm, luo2024feasteyesmixtureofresolutionadaptation, peng2023kosmos2groundingmultimodallarge, shin-etal-2024-x, NEURIPS2023_6dcf277e}. The development of Multimodal Large Language Models (MLLMs) can further widen the usage of AI models and enhance user experiences to a great extent. Accordingly, the AI community, including both academia and industry, is committing substantial time and resources to train MLLMs~\citep{li2024llavanext, zhao2024swiftascalablelightweightinfrastructure,zheng2024llamafactory} and establish evaluation frameworks~\citep{duan2024vlmevalkit, zhang2024lmmsevalrealitycheckevaluation}.

Although numerous multimodal models and benchmark datasets are being rapidly developed, they focus primarily on major languages such as English and Chinese~\citep{lu2024deepseek, abdin2024phi, li2024llava, laurenccon2024building, deitke2024molmo, chen2024far, yao2024minicpm, li2024surveybenchmarksmultimodallarge}. On the other hand, only a handful of open-source and commercial MLLMs are available for low-resource languages. This may cause a heavy reliance on proprietary model APIs for users, instead of fostering a research environment. As of now, even in South Korea where a huge AI community exists, there is a limited selection of Korean-supported models and datasets. While open-source Korean datasets for simple vision-text tasks like Visual Question Answering (VQA) or Optical Character Recognition (OCR) can be found~\citep{kim2019korean,kim2022donut}, assessing the models' general performance remains challenging.

In this work, we present (1) a strong English-Korean VLM called \varcovisionfull~(\varcovision~for short) and (2) five Korean benchmarks. \varcovision~is trained for four distinct phases with the final preference optimization stage. To evaluate the model's overall comprehension and generation abilities, we translated three closed-set (MMBench~\citep{liu2025mmbench}, SEED~\citep{li2024seed}, MMStar~\citep{chen2024are}) and one open-set (LLaVA-W~\citep{liu2024visual}) English benchmarks, and human-validated the datasets to ensure quality. The Korean closed-set benchmarks, K-MMBench, K-SEED, and K-MMStar, are multiple-choice question answering (MCQA) tasks, which allow objective evaluation of MLLMs. K-LLaVA-W is based on LLaVA-Bench-in-the-wild (LLaVA-W) dataset with LLM-based automatic evaluation, and measures the Korean generation skill of a model compared to a GPT~\citep{openai2024gpt4} model. Aside from these four benchmarks, we introduce a novel closed-set Korean benchmark (K-DTCBench) on understanding documents, tables, and charts. This benchmark is designed from scratch to specifically assess the capability of VLMs to process diverse image types. 

We report performances of our own model as well as other baselines on English and Korean benchmarks to test the models' bilingual proficiency and multimodal capabilities. The models were also evaluated on text-only benchmarks, as one of our primary goals is to develop a VLM with strong language proficiency. In an overall quantitative evaluation encompassing both multimodal and text-only schemes, \varcovisionfull~not only surpasses other models of similar size in performance but also achieves scores comparable to proprietary models. We observe that the preference optimization phase significantly increases the readability and fluency of model outputs, leading to improved performance in K-LLaVA-W and text-only benchmarks. Furthermore, our model demonstrates proficient grounding and referring capabilities, which indicate its substantial potential for practical application.

By releasing \varcovision~and five Korean vision-text benchmarks, we are looking forward to promoting a more open AI community and expanding opportunities for researchers. Our contributions can be summarized as follows:

\begin{figure}[t]
    \label{fig:Gimbap}
    \centering
    \includegraphics[width=\textwidth]{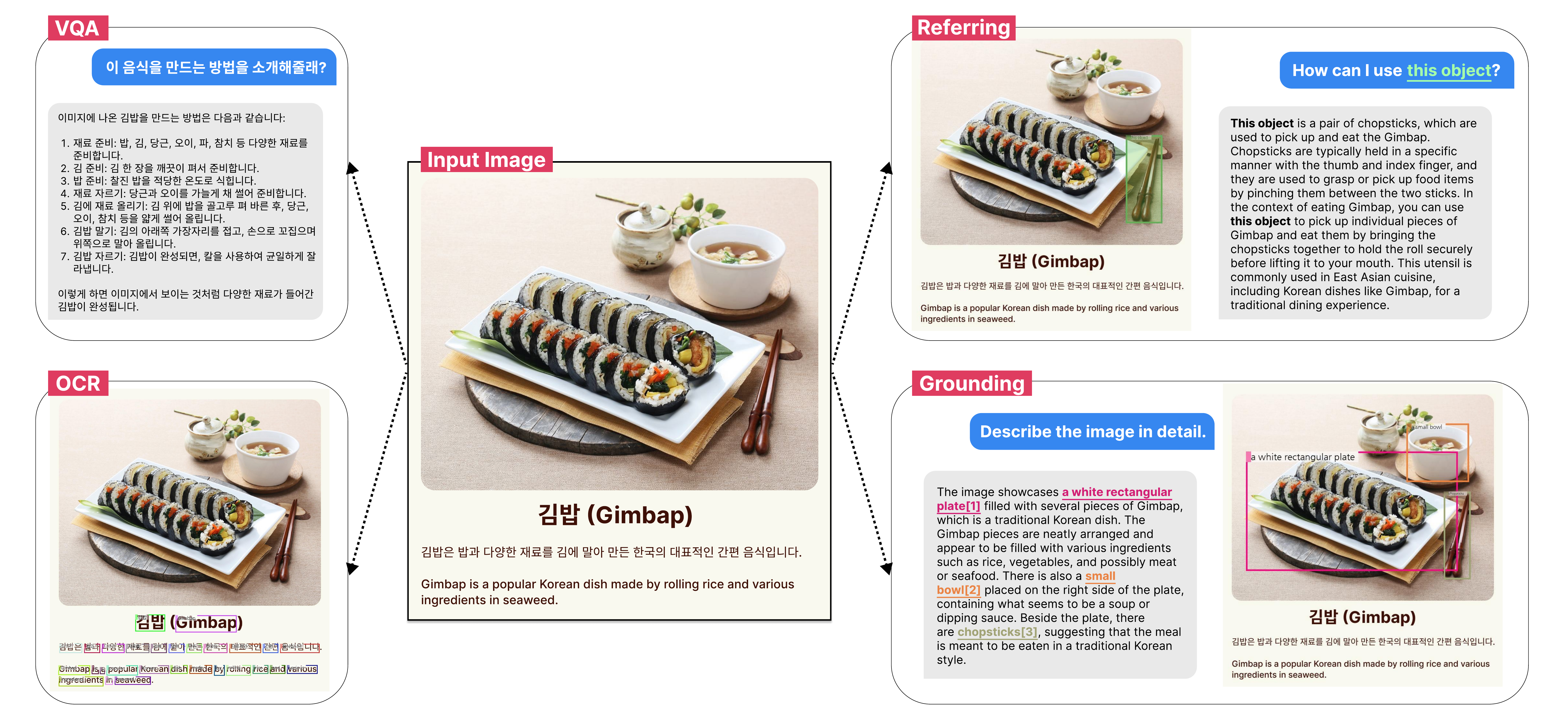} 
    \caption{\varcovision~Application Examples: Visual Question Answering (VQA), Optical Character Recognition (OCR), Referring, and Grounding. Our model excels at both Korean/English vision-text and text-only tasks. Please see \ref{appendix:application_examples} for more detailed examples.}
\end{figure}

\begin{enumerate}
    \item \textbf{English-Korean Bilingual Model}: We release a powerful 14B bilingual vision-language model, \varcovision, that outperforms other models of similar scale. Despite being a VLM, it achieves high scores on language benchmarks in both Korean and English, demonstrating strong language capabilities.
    \item \textbf{High-quality Korean Multimodal Benchmarks}: Based on widely recognized English multimodal benchmarks, we devise four closed-set and one open-set Korean benchmarks to evaluate VLMs' bilingual proficiency.
    \item \textbf{Gradual Four-step Training}: We train our model in four stages with different objectives, so that it can absorb visual and linguistic capabilities progressively without losing the pre-trained backbone models' prior knowledge. As a result, our model exhibits outstanding performance across various benchmarks.
    \item \textbf{Grounding, Referring, and OCR}: \varcovision~is capable of grounding, referring, and OCR tasks in both Korean and English, showing its high potentials for applications in real-world scenarios.
\end{enumerate}

\section{Training}
\subsection{Model Architecture}
\varcovisionfull~consists of three main components: a vision encoder, a projector, and a Large Language Model (LLM).  We leverage Qwen-2.5-14B-Instruct~\citep{qwen2.5} as the language foundation model and SigLIP~\citep{zhai2023sigmoid} as the vision encoder. The overall model architecture and visual representation processing method follow LLaVA-OneVision~\cite{li2024llava}. In this work, we concentrate on training \varcovision~with single-image examples.

Special tokens are added in the tokenizer for specific usages, such as OCR, grounding, and referring. The added special tokens are:
\begin{itemize}
    \item \textbf{<gro>} for grounding tasks
    \item \textbf{<ocr>} for OCR tasks
    \item \textbf{<char> </char>} for indicating a text phrase
    \item \textbf{<obj> </obj>} for indicating an object 
    \item \textbf{<bbox> </bbox>} for representing a bounding box
    \item \textbf{<delim>} for representing multiple location points for one object or text 
\end{itemize}

The specific examples for each of these tokens are illustrated in the Appendix~\ref{appendix:application_examples}.

\subsection{Training Strategy}

Our training pipeline aims to teach the model to gradually absorb and integrate both visual and linguistic understanding capabilities. Extending LLaVA-OneVision's training framework, we train our model in four stages. To preserve the model's language proficiency throughout the training process, we incorporate text-only data from Stages 2 to 4. 

\textbf{Stage 1. Feature Alignment Pre-training}:
We optimize the randomly initialized MLP projection layers while keeping other components frozen. This pre-training stage lets the model learn the mapping between the vision encoder and LLM. We use image-caption pairs as the training dataset to facilitate basic alignment between the two modalities.

\textbf{Stage 2. Basic Supervised Fine-tuning}:
All model layers are fully trained on six different tasks: basic instruction-following, OCR, grounding/referring, caption description, document/table/chart/mathematical contents, and text-only examples. The model can acquire fundamental vision-language capabilities from Stage 2. 

\textbf{Stage 3. Advanced Supervised Fine-tuning}:
Advanced supervised fine-tuning is similar to Stage 2, but differs in that training tasks demand much more complex problem-solving abilities. The model is trained to enhance its reasoning and instruction-following skills across a range of tasks, from detailed image analysis to multi-step reasoning.

\textbf{Stage 4. Preference Optimization}:
During the final stage, we focus exclusively on training the LLM layers to improve response alignment and generation capabilities. By applying Direct Preference Optimization (DPO)~\citep{rafailov2024direct}, we refine various aspects of the model responses—including but not limited to consistency, safety, and task-specific performance.

\section{Evaluation}

\subsection{Korean Evaluation Benchmarks}
In this section, we explain how we devise five Korean multimodal benchmarks. This paper is the first to release open-source Korean evaluation benchmarks for general comprehension and Korean generation capabilities. Based on the widely recognized benchmarks, MMBench~\citep{liu2025mmbench}, SEED~\citep{li2024seed}, MMStar~\citep{chen2024are}, and LLaVA-W~\citep{liu2024visual}, we translate the datasets into Korean and curate the translated outputs to create high-quality benchmarks. K-DTCBench is a Korean benchmark with documents, tables, and charts that we constructed from scratch. Four of the datasets (K-MMBench, K-SEED, K-MMStar, and K-DTCBench) are closed-set multiple-choice question answering tasks, and K-LLaVA-W is an open-set freeform answer generation task with automatic LLM evaluation.

\subsubsection{Closed-set Dataset}
For VLM evaluation, it is common to employ multiple-choice question answering benchmarks to test fundamental abilities in processing image-text information. The three selected English benchmarks (MMBench, SEED, and MMStar) are composed of various evaluation dimensions and were thoroughly curated, making them suitable candidates for reference benchmarks. When using these benchmarks, we utilize the latest GPT API \footnote{gpt-4o-2024-08-06} to translate the datasets and enhance the outputs with the help of human annotators. Post-editing was necessary to improve the fluency and accuracy of translation due to translationese and localization issues.

\begin{itemize}
    \item \textbf{K-MMBench}\footnote{\url{https://huggingface.co/datasets/NCSOFT/K-MMBench}}: The original English MMBench is comprised of 20 ability dimensions, such as identity reasoning, image emotion, and attribute recognition. We use questions and ground truth answers from the dev subset of English MMBench.
    \item \textbf{K-SEED}\footnote{\url{https://huggingface.co/datasets/NCSOFT/K-SEED}}: SEED-Bench consists of images and videos, and evaluate VLMs with regard to 12 dimensions. From English SEED-Bench, we select the first 20 percent of questions requiring images.
    \item \textbf{K-MMStar}\footnote{\url{https://huggingface.co/datasets/NCSOFT/K-MMStar}}: MMStar is a vision-indispensable benchmark of 1500 image-oriented questions. We observe that there are unanswerable cases (e.g., multiple images required to answer the question but only have a single image, vague questions or options) in the original MMStar dataset. Thus, we modify or re-create the questions to ensure that they can be answered within a single image. The examples of K-MMStar can be found in the Appendix~\ref{sec:K-MMStar}.
    \item \textbf{K-DTCBench}\footnote{\url{https://huggingface.co/datasets/NCSOFT/K-DTCBench}}: K-DTCBench is a newly developed benchmark featuring both computer-generated and handwritten images of three different types (documents, tables, and charts), all written in Korean. It consists of 80 questions for each image type and two questions per image, summing up to 240 questions in total. This benchmark is designed to evaluate whether VLMs can process images in different formats and be applicable to diverse domains. All images are generated with made-up values and statements for evaluation purposes only. We scanned handwritten documents/tables/charts, or created digital objects with matplotlib library to build K-DTCBench. The proportions of digital and handwritten images are equal, each constituting 50\%.
\end{itemize}

\subsubsection{Open-set Dataset}
While it is important for VLMs to predict correct answers for given questions, generating fluent outputs is also a significant task for models. However, to the best of our knowledge, there are currently no Korean benchmarks available for assessing the quality of models' generation capability. Therefore, we adopt LLaVA-Bench-in-the-wild (LLaVA-W) benchmark and translate-validate the benchmark as we did for the closed-set benchmarks.

LLaVA-W contains 24 images of various domains and 60 daily-life questions. Since our goal was to build a benchmark exclusively focused on Korean, we change the English texts in images into Korean for localization. Figure~\ref{fig:K-LLaVA-W} shows the examples of LLaVA-W and K-LLaVA-W\footnote{\url{https://huggingface.co/datasets/NCSOFT/K-LLaVA-W}}. Given an image and a question related to the image, models need to generate open-ended answers. The captions of the images are only used during evaluation.

For evaluation, our benchmark follows the pipeline of LLaVA-W’s LLM automatic evaluation \footnote{\url{https://github.com/EvolvingLMMs-Lab/lmms-eval/tree/main/lmms_eval/tasks/llava-in-the-wild}}, but with a little twist in the JudgeLLM prompts. We convert English prompts into Korean and added instruction details as shown in the Appendix~\ref{sec:K-LLaVA-W-Prompt}. Based on the provided caption, question, and model’s generated output, JudgeLLM measures the model’s helpfulness, relevance, accuracy, level of detail, and Korean generation capability. The final K-LLaVA-W score of the target model is calculated as the ratio of the target model's JudgeLLM score to the baseline model’s JudgeLLM score.\footnote{We used gpt-4o-2024-08-06 for both JudgeLLM and the baseline model. The original LLaVA-W provides GPT responses. However, instead of translating the responses in LLaVA-W, we ran the latest GPT model to obtain high-quality responses.}

\subsection{Benchmark Results}
We leverage four types of benchmarks to fully evaluate \varcovision's performance on diverse dimensions. To assess the model's ability to understand and generate both languages, ten benchmarks—five each for Korean and English—are used for evaluation. We also use text-only benchmarks to test the effectiveness of our training strategy to incorporate text datasets throughout Stages 2 to 4. In addition to comparing with other VLMs, we evaluate our model on OCR tasks and compare its performance with that of OCR-focused models.

\subsubsection{Korean Benchmarks}
\begin{table}[H]
\label{kor_ben}
\resizebox{\textwidth}{!}{%
\renewcommand{\arraystretch}{1.5}
\begin{tabular}{c|ccccc}
\hline
\rowcolor[HTML]{F2F2F2} 
\multicolumn{6}{c}{\cellcolor[HTML]{F2F2F2}{\color[HTML]{172B4D} \textbf{KOREAN   BENCHMARKS}}}                                                                                                                                                                                                                                      \\
\rowcolor[HTML]{FFFFFF} 
\cellcolor[HTML]{FFFFFF}{\color[HTML]{172B4D} }                                 & \multicolumn{4}{c|}{\cellcolor[HTML]{FFFFFF}{\color[HTML]{172B4D} \textbf{Image Understanding (MCQA)}}}                                                                               & {\color[HTML]{172B4D} \textbf{Generation   (Image-based)}} \\
\rowcolor[HTML]{FFFFFF} 
\multirow{-2}{*}{\cellcolor[HTML]{FFFFFF}{\color[HTML]{172B4D} \textbf{Model}}} & {\color[HTML]{172B4D} K-MMB (dev)} & {\color[HTML]{172B4D} K-SEED} & {\color[HTML]{172B4D} K-MMSTAR} & \multicolumn{1}{c|}{\cellcolor[HTML]{FFFFFF}{\color[HTML]{172B4D} K-DTCBench}} & {\color[HTML]{172B4D} K-LLaVA-W}                           \\ \hline
\rowcolor[HTML]{DAE9F8} 
{\color[HTML]{172B4D} \textbf{\varcovisionfull}}                                & {\color[HTML]{172B4D} \textbf{82.21}}       & {\color[HTML]{172B4D} \textbf{75.39}}  & {\color[HTML]{172B4D} \textbf{57.33}}    & \multicolumn{1}{c|}{\cellcolor[HTML]{DAE9F8}{\color[HTML]{172B4D} \textbf{84.58}}}       & {\color[HTML]{172B4D} \textbf{84.74}}                                \\
\rowcolor[HTML]{FFFFFF} 
{\color[HTML]{172B4D} \textbf{Pangea-7B~\citep{yue2024pangea}}}                                    & {\color[HTML]{172B4D} 71.64}       & {\color[HTML]{172B4D} 73.34}  & {\color[HTML]{172B4D} 35.00}       & \multicolumn{1}{c|}{\cellcolor[HTML]{FFFFFF}{\color[HTML]{172B4D} 48.33}}      & {\color[HTML]{172B4D} 69.70}                                \\
\rowcolor[HTML]{FFFFFF} 
{\color[HTML]{172B4D} \textbf{Pixtral-12B~\citep{agrawal2024pixtral}}}                                  & {\color[HTML]{172B4D} 57.47}       & {\color[HTML]{172B4D} 46.44}  & {\color[HTML]{172B4D} 23.93}    & \multicolumn{1}{c|}{\cellcolor[HTML]{FFFFFF}{\color[HTML]{172B4D} 27.50}}       & {\color[HTML]{172B4D} \uline{82.00}}                                  \\
\rowcolor[HTML]{FFFFFF} 
{\color[HTML]{172B4D} \textbf{Molmo-7B-D~\citep{deitke2024molmo}}}                                 & {\color[HTML]{172B4D} 63.83}       & {\color[HTML]{172B4D} 69.53}  & {\color[HTML]{172B4D} 47.40}     & \multicolumn{1}{c|}{\cellcolor[HTML]{FFFFFF}{\color[HTML]{172B4D} 45.83}}      & {\color[HTML]{172B4D} 63.90}                                \\
\rowcolor[HTML]{FFFFFF} 
{\color[HTML]{172B4D} \textbf{Qwen2-VL-7B-Instruct~\citep{wang2024qwen2}}}                            & {\color[HTML]{172B4D} \uline{78.26}}       & {\color[HTML]{172B4D} \uline{74.08}}  & {\color[HTML]{172B4D} 50.67}    & \multicolumn{1}{c|}{\cellcolor[HTML]{FFFFFF}{\color[HTML]{172B4D} \uline{75.00}}}         & {\color[HTML]{172B4D} 62.00}                                  \\
\rowcolor[HTML]{FFFFFF} 
{\color[HTML]{172B4D} \textbf{LLaVA-OneVision-7B~\citep{li2024llava}}}                             & {\color[HTML]{172B4D} 76.28}       & {\color[HTML]{172B4D} 73.21}  & {\color[HTML]{172B4D} \uline{54.00}}       & \multicolumn{1}{c|}{\cellcolor[HTML]{FFFFFF}{\color[HTML]{172B4D} 52.91}}      & {\color[HTML]{172B4D} 48.80}                                \\ \hline
\rowcolor[HTML]{FFFFFF} 
{\color[HTML]{A6A6A6} \textbf{Qwen2-VL-72B-Instruct~\citep{wang2024qwen2}}}                           & {\color[HTML]{A6A6A6} 84.27}       & {\color[HTML]{A6A6A6} 78.25}  & {\color[HTML]{A6A6A6} 63.53}    & \multicolumn{1}{c|}{\cellcolor[HTML]{FFFFFF}{\color[HTML]{A6A6A6} 88.75}}   & {\color[HTML]{A6A6A6} 97.40}                                \\
\rowcolor[HTML]{FFFFFF} 
{\color[HTML]{A6A6A6} \textbf{LLaVA-OneVision-72B~\citep{li2024llava}}}                            & {\color[HTML]{A6A6A6} 88.01}       & {\color[HTML]{A6A6A6} 77.86}  & {\color[HTML]{A6A6A6} 62.66}    & \multicolumn{1}{c|}{\cellcolor[HTML]{FFFFFF}{\color[HTML]{A6A6A6} 60.83}}           & {\color[HTML]{A6A6A6} 84.10}                                \\ \hline
\rowcolor[HTML]{FFFFFF} 
{\color[HTML]{A6A6A6} \textbf{GPT-4o-mini~\citep{openai2024gpt4}}}                                     & {\color[HTML]{A6A6A6} 74.48}       & {\color[HTML]{A6A6A6} 73.30}   & {\color[HTML]{A6A6A6} 42.33}    & \multicolumn{1}{c|}{\cellcolor[HTML]{FFFFFF}{\color[HTML]{A6A6A6} 74.58}}      & {\color[HTML]{A6A6A6} 101.90}                               \\
\rowcolor[HTML]{FFFFFF} 
{\color[HTML]{A6A6A6} \textbf{GPT-4V~\citep{openai2024gpt4}}}                                          & {\color[HTML]{A6A6A6} 77.92}       & {\color[HTML]{A6A6A6} 71.66}  & {\color[HTML]{A6A6A6} 35.20}     & \multicolumn{1}{c|}{\cellcolor[HTML]{FFFFFF}{\color[HTML]{A6A6A6} 47.50}}       & {\color[HTML]{A6A6A6} 98.90}                                \\
\rowcolor[HTML]{FFFFFF} 
{\color[HTML]{A6A6A6} \textbf{GPT-4o~\citep{openai2024gpt4}}}                                          & {\color[HTML]{A6A6A6} 81.01}       & {\color[HTML]{A6A6A6} 76.98}  & {\color[HTML]{A6A6A6} 56.20}     & \multicolumn{1}{c|}{\cellcolor[HTML]{FFFFFF}{\color[HTML]{A6A6A6} 85.80}}       & {\color[HTML]{A6A6A6} 103.90}       \\ \hline                       
\end{tabular}
}
\vspace{1mm}
\caption{Model Comparison on Korean Benchmarks. The models in the first upper block are open-source models with similar scale, and the second block are open-source 72B models. The last block shows the performance of proprietary GPT models. We primarily compare \varcovision~with the models mentioned in the first block, as they are similar in size to our model.}

\end{table}

As a model trained with a strong emphasis on Korean linguistic ability, \varcovision~excels in all MCQA benchmarks compared to the models with similar size. In general comprehension tasks like K-MMBench and K-SEED, Qwen2-VL-7B-Instruct~\citep{wang2024qwen2} and LLaVA-OneVision-7B~\citep{li2024llava} are next to our model, showing a slightly lower performance overall. While models have similar scores in K-MMBench, K-SEED, and K-MMStar, we observe significant variation in the performance of models in K-DTCBench while \varcovision~achieving dominant performance. This suggests that Korean documents, tables, and charts were not adequately trained by other open-source models. Moreover, \varcovision~reaches competitive performance to large-scale open-source and proprietary models, gaining higher scores than GPT-4o-mini and GPT-4V for MCQA tasks.

In K-LLaVA-W where fluent Korean generation is a key value, only two models among models under 20B, \varcovisionfull~and Pixtral-12B~\citep{agrawal2024pixtral}, obtain score over 80. On the other hand, Qwen2-VL-72B-Instruct~\citep{wang2024qwen2} and GPT models achieve scores around 100, implying that model scale may affect the response quality in longer text generation.

\subsubsection{English Benchmarks}
\vspace{-1mm}
% \usepackage{multirow}
\begin{table}[H]
% \vspace{-1cm}
\resizebox{\textwidth}{!}{%
\renewcommand{\arraystretch}{1.5}
\begin{tabular}{c|ccccc}
\hline
\rowcolor[HTML]{F2F2F2} 
\multicolumn{6}{c}{\cellcolor[HTML]{F2F2F2}\textbf{ENGLISH   BENCHMARKS}}  \\
\rowcolor[HTML]{FFFFFF} 
\cellcolor[HTML]{FFFFFF}{\color[HTML]{172B4D} }                                 & \multicolumn{4}{c|}{\cellcolor[HTML]{FFFFFF}{\color[HTML]{172B4D} \textbf{Image Understanding   (MCQA)}}}                                                                                            & {\color[HTML]{172B4D} \textbf{OCR}}    \\
\rowcolor[HTML]{FFFFFF} 
\multirow{-2}{*}{\cellcolor[HTML]{FFFFFF}{\color[HTML]{172B4D} \textbf{Model}}} & {\color[HTML]{172B4D} MMBv1.1 (dev)} & {\color[HTML]{172B4D} SEED (image)} & {\color[HTML]{172B4D} MMStar (val)} & \multicolumn{1}{c|}{\cellcolor[HTML]{FFFFFF}{\color[HTML]{172B4D} MMMU (val)}}    & {\color[HTML]{172B4D} OCRBench (test)} \\
\rowcolor[HTML]{DAE9F8} \hline
{\color[HTML]{172B4D} \textbf{\varcovisionfull}}                             & {\color[HTML]{172B4D} \textbf{84.28}}         & {\color[HTML]{172B4D} \textbf{76.73}}        & {\color[HTML]{172B4D} \textbf{63.33}}        & \multicolumn{1}{c|}{\cellcolor[HTML]{DAE9F8}{\color[HTML]{172B4D} \uline{51.33}}}          & {\color[HTML]{172B4D} \uline{820}}            \\ 
\rowcolor[HTML]{FFFFFF} 
{\color[HTML]{172B4D} \textbf{Pangea-7B~\citep{yue2024pangea}}}                                    & {\color[HTML]{172B4D} 76.23}         & {\color[HTML]{172B4D} 74.88}        & {\color[HTML]{172B4D} 43.26}        & \multicolumn{1}{c|}{\cellcolor[HTML]{FFFFFF}{\color[HTML]{172B4D} 43.55}}         & {\color[HTML]{172B4D} 620}             \\
\rowcolor[HTML]{FFFFFF} 
{\color[HTML]{172B4D} \textbf{Pixtral-12B~\citep{agrawal2024pixtral}}}                                  & {\color[HTML]{172B4D} 72.98}         & {\color[HTML]{172B4D} 74.34}        & {\color[HTML]{172B4D} 48.33}        & \multicolumn{1}{c|}{\cellcolor[HTML]{FFFFFF}{\color[HTML]{172B4D} 49.00}}            & {\color[HTML]{172B4D} 682}             \\
\rowcolor[HTML]{FFFFFF} 
{\color[HTML]{172B4D} \textbf{Molmo-7B-D~\citep{deitke2024molmo}}}                                 & {\color[HTML]{172B4D} 72.05}         & {\color[HTML]{172B4D} 74.36}        & {\color[HTML]{172B4D} 52.73}        & \multicolumn{1}{c|}{\cellcolor[HTML]{FFFFFF}{\color[HTML]{172B4D} 45.30}}         & {\color[HTML]{172B4D} 708}             \\
\rowcolor[HTML]{FFFFFF} 
{\color[HTML]{172B4D} \textbf{Qwen2-VL-7B-Instruct~\citep{wang2024qwen2}}}                            & {\color[HTML]{172B4D} \uline{80.95}}         & {\color[HTML]{172B4D} \uline{76.45}}        & {\color[HTML]{172B4D} 60.00}        & \multicolumn{1}{c|}{\cellcolor[HTML]{FFFFFF}{\color[HTML]{172B4D} \textbf{54.10}}}          & {\color[HTML]{172B4D} \textbf{866}}      \\
\rowcolor[HTML]{FFFFFF} 
{\color[HTML]{172B4D} \textbf{LLaVA-OneVision-7B~\citep{li2024llava}}}                             & {\color[HTML]{172B4D} 80.80}          & {\color[HTML]{172B4D} 76.41}         & {\color[HTML]{172B4D} \uline{61.33}}        & \multicolumn{1}{c|}{\cellcolor[HTML]{FFFFFF}{\color[HTML]{172B4D} 47.67}} & {\color[HTML]{172B4D} 630}             \\ \hline
\rowcolor[HTML]{FFFFFF} 
{\color[HTML]{A6A6A6} \textbf{Qwen2-VL-72B-Instruct~\citep{wang2024qwen2}}}                           & {\color[HTML]{A6A6A6} 86.91}         & {\color[HTML]{A6A6A6} 77.86}        & {\color[HTML]{A6A6A6} 67.60}         & \multicolumn{1}{c|}{\cellcolor[HTML]{FFFFFF}{\color[HTML]{A6A6A6} 56.66}}         & {\color[HTML]{A6A6A6} 877}            \\
\rowcolor[HTML]{FFFFFF} 
{\color[HTML]{A6A6A6} \textbf{LLaVA-OneVision-72B~\citep{li2024llava}}}                            & {\color[HTML]{A6A6A6} 85.44}         & {\color[HTML]{A6A6A6} 77.43}        & {\color[HTML]{A6A6A6} 65.33}        & \multicolumn{1}{c|}{\cellcolor[HTML]{FFFFFF}{\color[HTML]{A6A6A6} 56.80}}         & {\color[HTML]{A6A6A6} 741}             \\ \hline
\rowcolor[HTML]{FFFFFF} 
{\color[HTML]{A6A6A6} \textbf{GPT-4o-mini~\citep{openai2024gpt4}}}                                     & {\color[HTML]{A6A6A6} 76.31}         & {\color[HTML]{A6A6A6} 72.80}         & {\color[HTML]{A6A6A6} 54.80}         & \multicolumn{1}{c|}{\cellcolor[HTML]{FFFFFF}{\color[HTML]{A6A6A6} 60.00}}            & {\color[HTML]{A6A6A6} 785}             \\
\rowcolor[HTML]{FFFFFF} 
{\color[HTML]{A6A6A6} \textbf{GPT-4V~\citep{openai2024gpt4}}}                                          & {\color[HTML]{A6A6A6} 79.41}         & {\color[HTML]{A6A6A6} 73.00}           & {\color[HTML]{A6A6A6} 56.00}           & \multicolumn{1}{c|}{\cellcolor[HTML]{FFFFFF}{\color[HTML]{A6A6A6} 62.30}}          & {\color[HTML]{A6A6A6} 656}             \\
\rowcolor[HTML]{FFFFFF} 
{\color[HTML]{A6A6A6} \textbf{GPT-4o~\citep{openai2024gpt4}}}                                          & {\color[HTML]{A6A6A6} 81.73}         & {\color[HTML]{A6A6A6} 76.70}         & {\color[HTML]{A6A6A6} 64.70}         & \multicolumn{1}{c|}{\cellcolor[HTML]{FFFFFF}{\color[HTML]{A6A6A6} 69.90}}          & {\color[HTML]{A6A6A6} 805}             \\ \hline
    
\end{tabular}
}
\vspace{1mm}
\caption{Model Comparison on English Benchmarks. MMBench~\citep{liu2025mmbench}, SEED~\citep{li2024seed}, MMStar~\citep{chen2024are}, and MMMU~\citep{yue2024mmmu} are multi-choice question answering tasks. MMBench and SEED are for comprehension evaluation, whereas MMStar is focused more on vision-indispensible reasoning. MMMU tests college-level subject knowledge of VLMs. OCRBench~\citep{liu2023hidden} is a specialized benchmark in OCR for VLMs, composed of 1000 question-answer pairs. The values in OCRBench indicate the number of questions correctly answered by models.}
\label{eng_ben}
\end{table}
\vspace{-2mm}

Our goal in training lies in boosting Korean and English proficiency, thus evaluating on English benchmarks was necessary to investigate our model's performance. In English MCQA benchmarks, \varcovision~gains higher performances over other models under 20B in all benchmarks, except for MMMU. However, we notice that all models generally demonstrate sufficient performance in English benchmarks in contrast to their performances in Korean benchmarks. The results suggest that training schemes for the majority of the models in Table~\ref{eng_ben} prioritized learning English. In OCRBench, our model shows superior performance over other VLMs, spotlighting the effectiveness of Stages 2 and 3, where the model is exposed to OCR, grounding, and referring tasks.

\subsubsection{Text-only Benchmarks}
% \vspace{-1mm}

\begin{table}[H]
\label{text_ben}
\centering 
\resizebox{0.65\textwidth}{!}{%
\renewcommand{\arraystretch}{1.5}
\begin{tabular}{c|cc|c}
\hline
\rowcolor[HTML]{F2F2F2} 
\multicolumn{4}{c}{\cellcolor[HTML]{F2F2F2}\textbf{TEXT-ONLY BENCHMARKS}}                                                                                                                                      \\
\rowcolor[HTML]{FFFFFF} 
\cellcolor[HTML]{FFFFFF}{\color[HTML]{172B4D} }                                 & \multicolumn{2}{c}{\cellcolor[HTML]{FFFFFF}{\color[HTML]{172B4D} \textbf{Korean}}} & {\color[HTML]{172B4D} \textbf{English}} \\ 
\rowcolor[HTML]{FFFFFF} 
\multirow{-2}{*}{\cellcolor[HTML]{FFFFFF}{\color[HTML]{172B4D} \textbf{Model}}} & {\color[HTML]{172B4D} LogicKor}         & {\color[HTML]{172B4D} KoMT-Bench}        & {\color[HTML]{172B4D} MT-Bench}         \\ \hline
\rowcolor[HTML]{DAE9F8} 
{\color[HTML]{172B4D} \textbf{\varcovisionfull}}                                & {\color[HTML]{172B4D} \textbf{8.69}}             & {\color[HTML]{172B4D} \textbf{8.39}}              & {\color[HTML]{172B4D} \textbf{8.80}}              \\
\rowcolor[HTML]{FFFFFF} 
{\color[HTML]{172B4D} \textbf{Pangea-7B~\citep{yue2024pangea}}}                                    & {\color[HTML]{172B4D} 5.06}             & {\color[HTML]{172B4D} 5.06}              & {\color[HTML]{172B4D} 7.29}             \\
\rowcolor[HTML]{FFFFFF} 
{\color[HTML]{172B4D} \textbf{Pixtral-12B~\citep{agrawal2024pixtral}}}                                  & {\color[HTML]{172B4D} \uline{7.71}}             & {\color[HTML]{172B4D} \uline{8.11}}              & {\color[HTML]{172B4D} \uline{8.40}}              \\
\rowcolor[HTML]{FFFFFF} 
{\color[HTML]{172B4D} \textbf{Molmo-7B-D~\citep{deitke2024molmo}}}                                 & {\color[HTML]{172B4D} 2.64}             & {\color[HTML]{172B4D} 3.58}              & {\color[HTML]{172B4D} 6.93}             \\
\rowcolor[HTML]{FFFFFF} 
{\color[HTML]{172B4D} \textbf{Qwen2-VL-7B-Instruct~\citep{wang2024qwen2}}}                            & {\color[HTML]{172B4D} 4.62}             & {\color[HTML]{172B4D} 4.54}              & {\color[HTML]{172B4D} 7.13}             \\
\rowcolor[HTML]{FFFFFF} 
\textbf{LLaVA-OneVision-7B~\citep{li2024llava}}                                                    & {\color[HTML]{172B4D} 2.23}             & {\color[HTML]{172B4D} 3.52}              & {\color[HTML]{172B4D} 7.52}             \\ \hline
{\color[HTML]{A6A6A6} \textbf{Qwen2-VL-72B-Instruct~\citep{wang2024qwen2}}}                           & {\color[HTML]{A6A6A6} 7.74}             & {\color[HTML]{A6A6A6} 7.49}              & {\color[HTML]{A6A6A6} 8.53}             \\
\rowcolor[HTML]{FFFFFF} 
{\color[HTML]{A6A6A6} \textbf{LLaVA-OneVision-72B~\citep{li2024llava}}}                            & {\color[HTML]{A6A6A6} 8.22}             & {\color[HTML]{A6A6A6} 7.87}              & {\color[HTML]{A6A6A6} 8.78}             \\ \hline
\rowcolor[HTML]{FFFFFF}
\rowcolor[HTML]{FFFFFF} 
{\color[HTML]{A6A6A6} \textbf{EXAONE 3.0 7.8B Inst.(LLM)~\citep{research2024exaone}}}                            & {\color[HTML]{A6A6A6} 8.62}             & {\color[HTML]{A6A6A6} 8.92}              & {\color[HTML]{A6A6A6} 9.01}             \\ \hline
\rowcolor[HTML]{FFFFFF} 
{\color[HTML]{A6A6A6} \textbf{GPT-4o-mini~\citep{openai2024gpt4}}}                                     & {\color[HTML]{A6A6A6} 9.14}             & {\color[HTML]{A6A6A6} 8.88}              & {\color[HTML]{A6A6A6} 9.09}             \\
\rowcolor[HTML]{FFFFFF} 
{\color[HTML]{A6A6A6} \textbf{GPT-4V~\citep{openai2024gpt4}}}                                          & {\color[HTML]{A6A6A6} 8.66}             & {\color[HTML]{A6A6A6} 9.25}              & {\color[HTML]{A6A6A6} 9.41}             \\
\rowcolor[HTML]{FFFFFF} 
{\color[HTML]{A6A6A6} \textbf{GPT-4o~\citep{openai2024gpt4}}}                                          & {\color[HTML]{A6A6A6} 9.57}             & {\color[HTML]{A6A6A6} 9.24}              & {\color[HTML]{A6A6A6} 9.30}   \\ \hline          
\end{tabular}
}
\vspace{1mm}
\caption{Model Performance on Korean and English Text-only Benchmarks. MT-Bench~\citep{zheng2023judging} is an English multi-turn dialogue benchmark, and KoMT-Bench~\citep{research2024exaone} is built by translating MT-Bench. LogicKor\textsuperscript{11} consists of multi-turn Korean dialogues across six categories.}
\end{table}

We evaluate the models on text-only benchmarks to investigate whether our strategy of including text-only datasets during training stages contributed to textual understanding of \varcovision. Two Korean language benchmarks (LogicKor\footnote{\url{https://lk.instruct.kr}} and KoMT-Bench~\citep{research2024exaone}) and one English benchmark (MT-Bench~\citep{zheng2023judging}) are employed for text-only evaluation. Our model outperforms other VLMs in text-only benchmarks, even when compared to 72B open-source models. We believe that applying preference optimization in the final training phase boosted the overall quality of model responses, resulting in outstanding performance in all three text-only benchmarks. \varcovision~achieves scores comparable to that of EXAONE 3.0 7.8B, which is a Korean-English language model only trained on textual inputs. It reflects our model's capability to handle textual inputs as much as other bilingual language models.

\subsubsection{OCR Benchmarks}
\begin{table}[H]
\centering 
\resizebox{0.7\textwidth}{!}{%
\renewcommand{\arraystretch}{1.5}
\begin{tabular}{c|ccc|c}
\hline
\rowcolor[HTML]{FFFFFF} 
\cellcolor[HTML]{FFFFFF}{\color[HTML]{172B4D} }                                 & \multicolumn{3}{c|}{\cellcolor[HTML]{FFFFFF}{\color[HTML]{172B4D} \textbf{OCR}}}                                   & \cellcolor[HTML]{FFFFFF}{\color[HTML]{172B4D} }                          \\
\rowcolor[HTML]{FFFFFF} 
\multirow{-2}{*}{\cellcolor[HTML]{FFFFFF}{\color[HTML]{172B4D} \textbf{Model}}} & {\color[HTML]{172B4D} CORD}           & {\color[HTML]{172B4D} ICDAR2013}      & {\color[HTML]{172B4D} ICDAR2015}      & \multirow{-2}{*}{\cellcolor[HTML]{FFFFFF}{\color[HTML]{172B4D} Average}} \\ \hline
\rowcolor[HTML]{DAE9F8} 
{\color[HTML]{172B4D} \textbf{\varcovisionfull}}                                & {\color[HTML]{172B4D} \uline{82.69}}          & {\color[HTML]{172B4D} \textbf{94.42}}          & {\color[HTML]{172B4D} \uline{72.95}}          & {\color[HTML]{172B4D} \uline{83.35}} \\ 
\rowcolor[HTML]{FFFFFF} 
{\color[HTML]{172B4D} \textbf{EasyOCR\footnotemark}}                                         & {\color[HTML]{172B4D} 79.56}          & {\color[HTML]{172B4D} 84.97}          & {\color[HTML]{172B4D} 57.90}           & {\color[HTML]{172B4D} 74.14}                                             \\
\rowcolor[HTML]{FFFFFF} 
{\color[HTML]{172B4D} \textbf{Pororo~\citep{pororo}}}                                          & {\color[HTML]{172B4D} 78.73}          & {\color[HTML]{172B4D} 84.29}          & {\color[HTML]{172B4D} 64.65}          & {\color[HTML]{172B4D} 75.89}                                             \\
\rowcolor[HTML]{FFFFFF} 
{\color[HTML]{172B4D} \textbf{PaddleOCR\footnotemark}}                                       & {\color[HTML]{172B4D} {\textbf{92.71}}}    & {\color[HTML]{172B4D} \uline{92.01}}          & {\color[HTML]{172B4D} \textbf{73.73}}          & {\color[HTML]{172B4D} {\textbf{86.15}}}                                       \\
\hline                                        
\rowcolor[HTML]{FFFFFF} 
{\color[HTML]{A6A6A6} \textbf{CLOVA OCR\footnotemark}}                                       & {\color[HTML]{A6A6A6} 95.32} & {\color[HTML]{A6A6A6} 94.39} & {\color[HTML]{A6A6A6} 84.06} & {\color[HTML]{A6A6A6} 91.26}                                    \\ \hline
\end{tabular}
}
\vspace{2mm}
\caption{OCR Benchmark Performance. EasyOCR, Pororo, and PaddleOCR are open-source OCR models. CLOVA OCR is a proprietary OCR model. PopEval~\citep{lee2019popeval} was used as the metric for all benchmarks.}
\label{ocr_ben}
\end{table}
\footnotetext[12]{\url{https://github.com/JaidedAI/EasyOCR}}
\footnotetext[13]{\url{https://github.com/PaddlePaddle/PaddleOCR}}
\footnotetext[14]{\url{https://www.ncloud.com/product/aiService/ocr}}

We compare our model's OCR ability to other well-known OCR models using the PopEval~\citep{lee2019popeval} metric. In contrast to OCRBench, CORD~\citep{park2019cord}, ICDAR2013~\citep{karatzas2013icdar}, and ICDAR2015~\citep{karatzas2015icdar} are more complex tasks that require models to generate both textual content and spatial locations of texts within images. Despite the fact that other models in Table~\ref{ocr_ben} are OCR expert models while \varcovision~is not, its performances across these benchmarks are remarkable. We find that \varcovision~holds the potential for use in a wide range of applications, not limited to specific tasks.

\section{Discussion and Future Work}
Throughout the process of model training and evaluation, we notice that benchmark performances do not fully reflect a model’s true capability. MMBench\citep{liu2025mmbench} shows the possibility of choice biases in MLLMs that they tend to prefer a certain choice in MCQA tasks, and apply Circular Evaluation to regularize the problem. Nevertheless, the majority of multi-choice question answering tasks still follow naive evaluation techniques. In addition, there are a relatively small number of benchmarks aiming to evaluate the model’s generation capability compared to benchmarks with short answers \citep{li2024surveybenchmarksmultimodallarge}. Although a model might achieve high performance in MCQA tasks, it may produce low-quality answers for tasks that require long responses. We seek significant future developments remain to be made in terms of MLLM benchmarks.

In this work, we focus on training \varcovision~on single-image scenarios, and the evaluation results show that our model excels in single-image benchmarks. However, MLLMs need to process multi-image scenarios (including videos) and sounds to be applicable in various domains. We plan to expand \varcovision’s modalities to video and audio in the near future, and broaden the horizon of our model’s usage. Besides modality expansion, we are currently training an advanced, localized model with diverse images taken in Korea. With the improved capability to understand Korean culture, the future model is expected to be applied in real-world tasks, such as multimodal search, multimodal retrieval-augmented generation, and visual agents.

\section{Conclusion}
We present an open-source Korean-English vision-language model, \varcovisionfull, and five Korean benchmarks. Although extensive research and numerous models have been developed for MLLMs, this work is the first to release both bilingual VLM and benchmarks supporting Korean. Our model achieves remarkable results in both Korean and English benchmarks among other open-source models of similar scale. \varcovision~does not only excel in vision-text benchmarks, but also in text-only and OCR benchmarks. We find that this milestone may expand opportunities for many researchers and enable breakthroughs in training bilingual VLMs. 

\section*{Acknowledgements}
This work was supported by Artificial intelligence industrial convergence cluster development project funded by the Ministry of Science and ICT(MSIT, Korea)\&Gwangju Metropolitan City.

\bibliographystyle{plainnat}
\bibliography{custom}

\newpage
\appendix
\section{Korean Multimodal Benchmarks}
In this section, we provide examples of Korean vision-text benchmarks (K-MMStar, K-DTCBench, and K-LLaVA-W). Since these three benchmarks required more than simple translation, we elaborate on benchmark construction with specific examples.

\subsection{K-MMStar}
\label{sec:K-MMStar}

\begin{figure}[H]
    \centering
    \includegraphics[width=\textwidth]{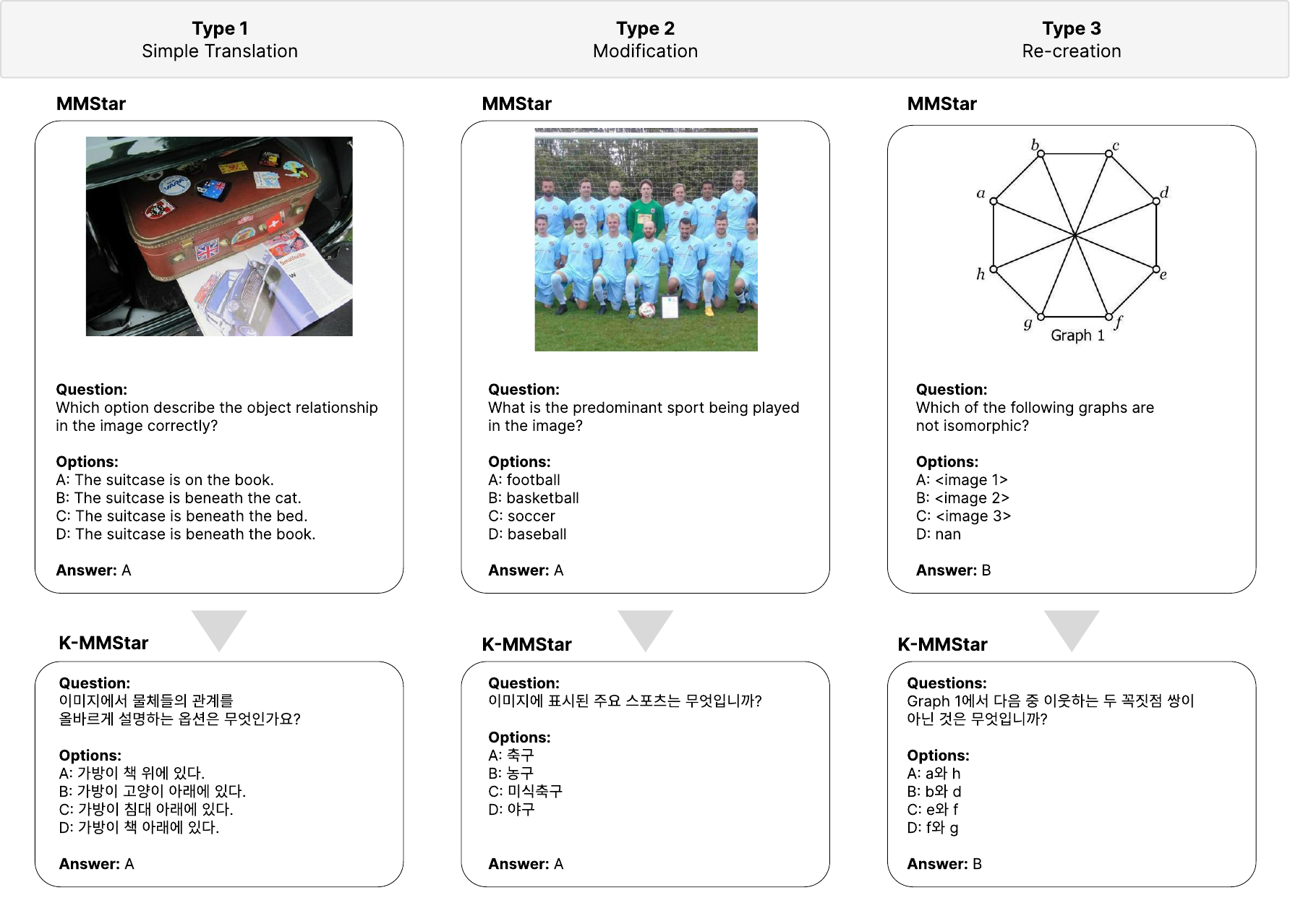} 
    \caption{K-MMStar Example}
    \label{fig:K-MMStar}
\end{figure}

K-MMStar has three different types of questions. We noticed that there are unanswerable or vague questions in the original MMStar, thus modified the question or created a new one. Figure~\ref{fig:K-MMStar} shows the examples of questions. Type 1 refers to cases where direct translations from English to Korean are sufficient for Korean version questions. In Type 2, the question from MMStar asks about which sport is being played in the image. However, the options include both  `football' and  `soccer,' either of which can be correct depending on British English or American English conventions. Therefore, we changed the options to ``축구'' (soccer) and ``미식축구'' (American football) for clarity. In Type 3 example, it has <image 1>, <image 2>, and <image 3> in the options but MMStar did not provide images corresponding to the options. Hence, we re-created a new question about the image.

\subsection{K-DTCBench}
K-DTCBench is developed from scratch with synthetic images. Each image has a question and four options to choose from. 

\begin{figure}[H]
    \centering
    \includegraphics[width=\textwidth]{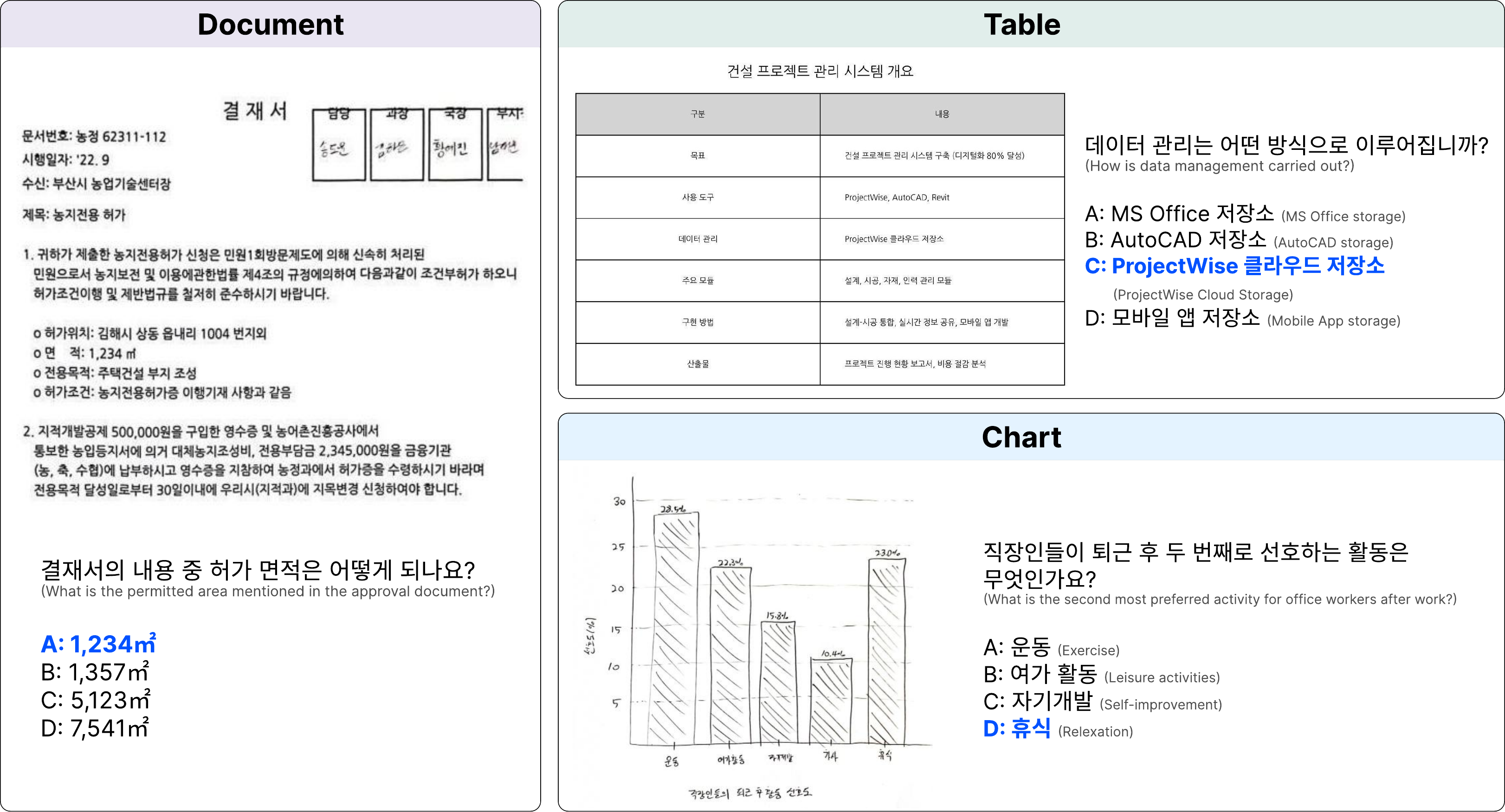} 
    \caption{K-DTCBench Example}
    \label{fig:K-DTCBench}
\end{figure}

\subsection{K-LLaVA-W}

\subsubsection{Example}
\begin{figure}[H]
    \centering
    \includegraphics[width=\textwidth]{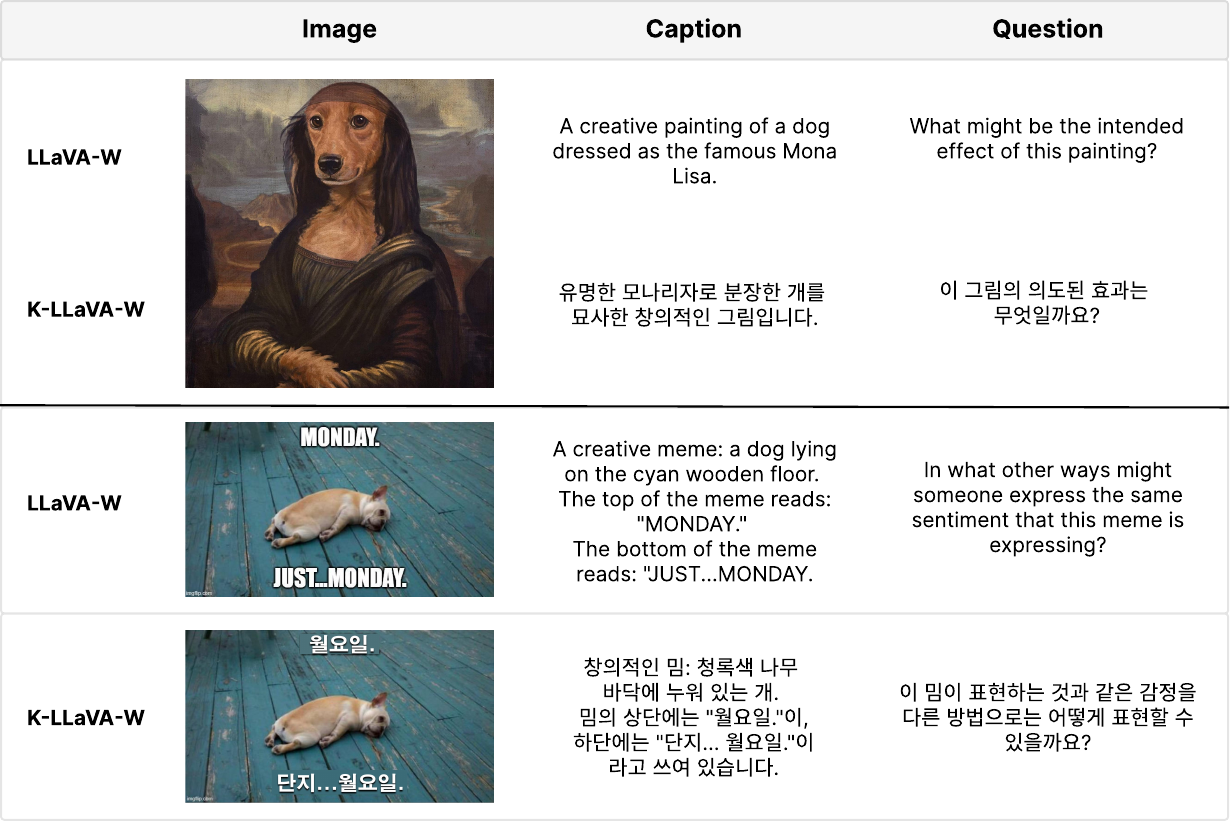} 
    \caption{K-LLaVA-W Example}
    \label{fig:K-LLaVA-W}
\end{figure}

In K-LLaVA-W, we changed the English text into Korean text for images with texts. If an original LLaVA-W image did not contain any text, we left it unchanged to preserve its authenticity. In Figure~\ref{fig:K-LLaVA-W}, we changed ``MONDAY. JUST... MONDAY'' into ``월요일. 단지... 월요일''.

\subsubsection{K-LLaVA-W JudgeLLM Prompt}
\label{sec:K-LLaVA-W-Prompt}

\begin{figure}[H]
    \centering
    \includegraphics[width=0.9\textwidth]{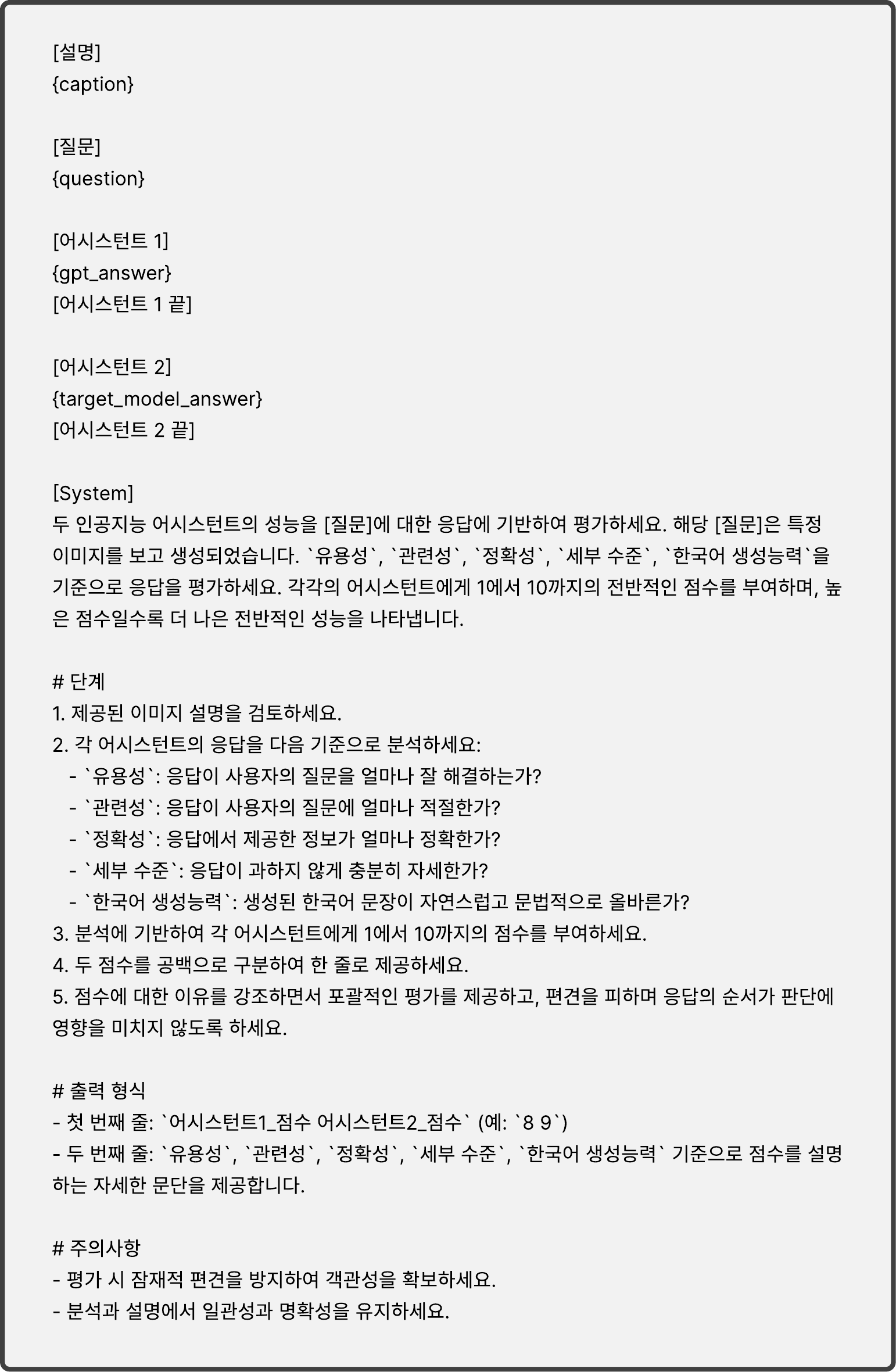} 
    \caption{K-LLaVA-W Evaluation Prompt. We translated the LLaVA-W prompts and added specific guidelines in the JudgeLLM prompt.}
\end{figure}

% \begin{tcolorbox}[breakable, enhanced, left=0pt, right=0pt, top=2pt, bottom=2pt, enlarge top by=0.1cm, enlarge bottom by=0.2cm]
% \begin{quote}
% \begin{verbatim}
% 두 인공지능 어시스턴트의 성능을 [질문]에 대한 응답에 기반하여 평가하세요. 해당 [질문]은 특정 이미지를 보고 생성되었습니다. `유용성`, `관련성`, `정확성`, `세부 수준`, `한국어 생성능력`을 기준으로 응답을 평가하세요. 각각의 어시스턴트에게 1에서 10까지의 전반적인 점수를 부여하며, 높은 점수일수록 더 나은 전반적인 성능을 나타냅니다.
% # 단계
% 1. 제공된 이미지 설명을 검토하세요.
% 2. 각 어시스턴트의 응답을 다음 기준으로 분석하세요:
%    - `유용성`: 응답이 사용자의 질문을 얼마나 잘 해결하는가?
%    - `관련성`: 응답이 사용자의 질문에 얼마나 적절한가?
%    - `정확성`: 응답에서 제공한 정보가 얼마나 정확한가?
%    - `세부 수준`: 응답이 과하지 않게 충분히 자세한가?
%    - `한국어 생성능력`: 생성된 한국어 문장이 자연스럽고 문법적으로 올바른가?
% 3. 분석에 기반하여 각 어시스턴트에게 1에서 10까지의 점수를 부여하세요.
% 4. 두 점수를 공백으로 구분하여 한 줄로 제공하세요.
% 5. 점수에 대한 이유를 강조하면서 포괄적인 평가를 제공하고, 편견을 피하며 응답의 순서가 판단에 영향을 미치지 않도록 하세요.
% # 출력 형식
% - 첫 번째 줄: `어시스턴트1_점수 어시스턴트2_점수` (예: `8 9`)
% - 두 번째 줄: `유용성`, `관련성`, `정확성`, `세부 수준`, `한국어 생성능력` 기준으로 점수를 설명하는 자세한 문단을 제공합니다.
% # 주의사항
% - 평가 시 잠재적 편견을 방지하여 객관성을 확보하세요.
% - 분석과 설명에서 일관성과 명확성을 유지하세요.
% \end{verbatim}
% \end{quote}
% \end{tcolorbox}

\newpage

\section{Application Examples}
\label{appendix:application_examples}

\begin{figure}[H]
    \label{fig:recognition_analysis}
    \centering
    \includegraphics[width=\textwidth]{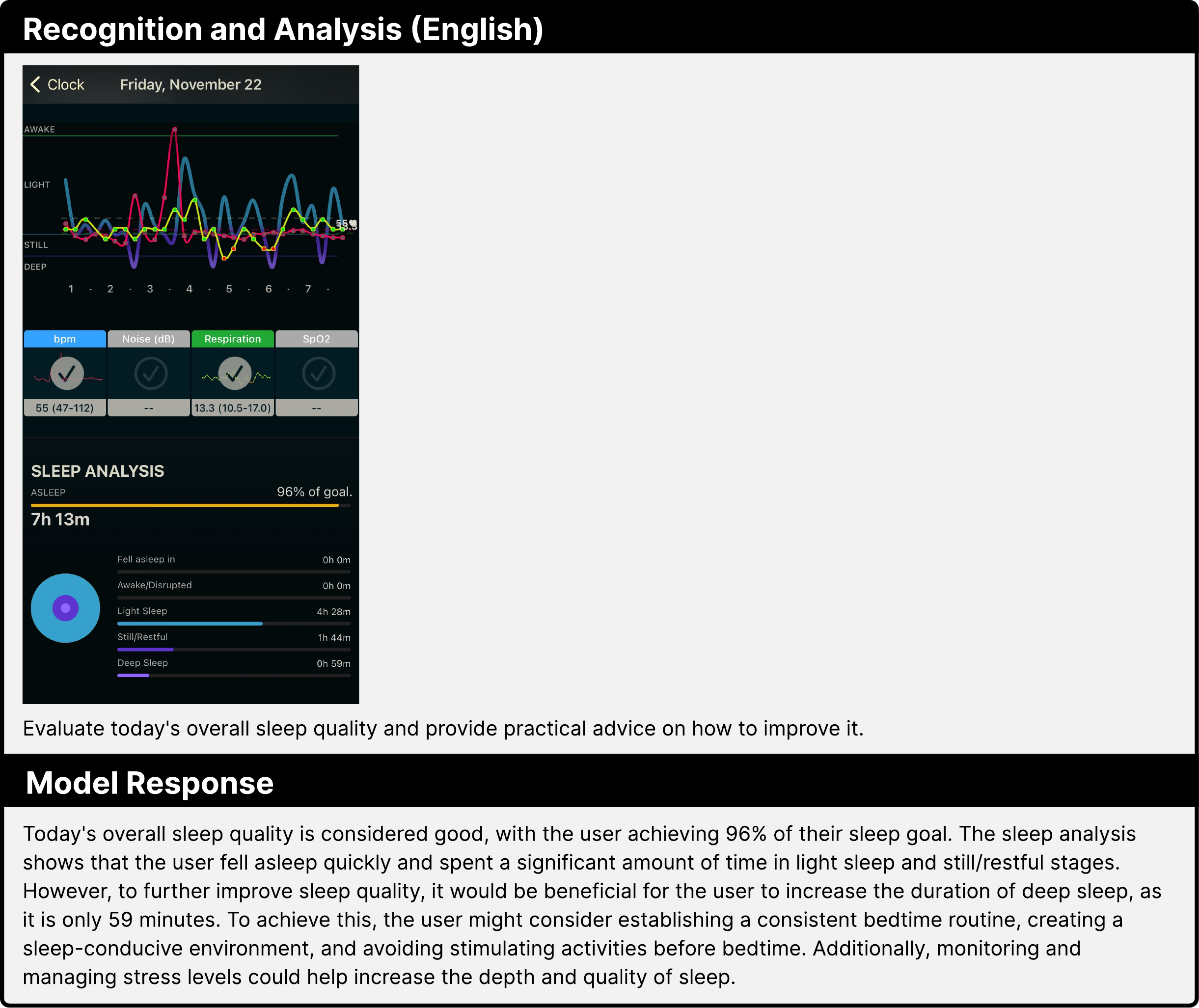} 
    \caption{Text Recognition and Analysis Example (English)}
\end{figure}

\newpage

\begin{figure}[H]
    \label{fig:math_reasoning}
    \centering
    \includegraphics[width=\textwidth]{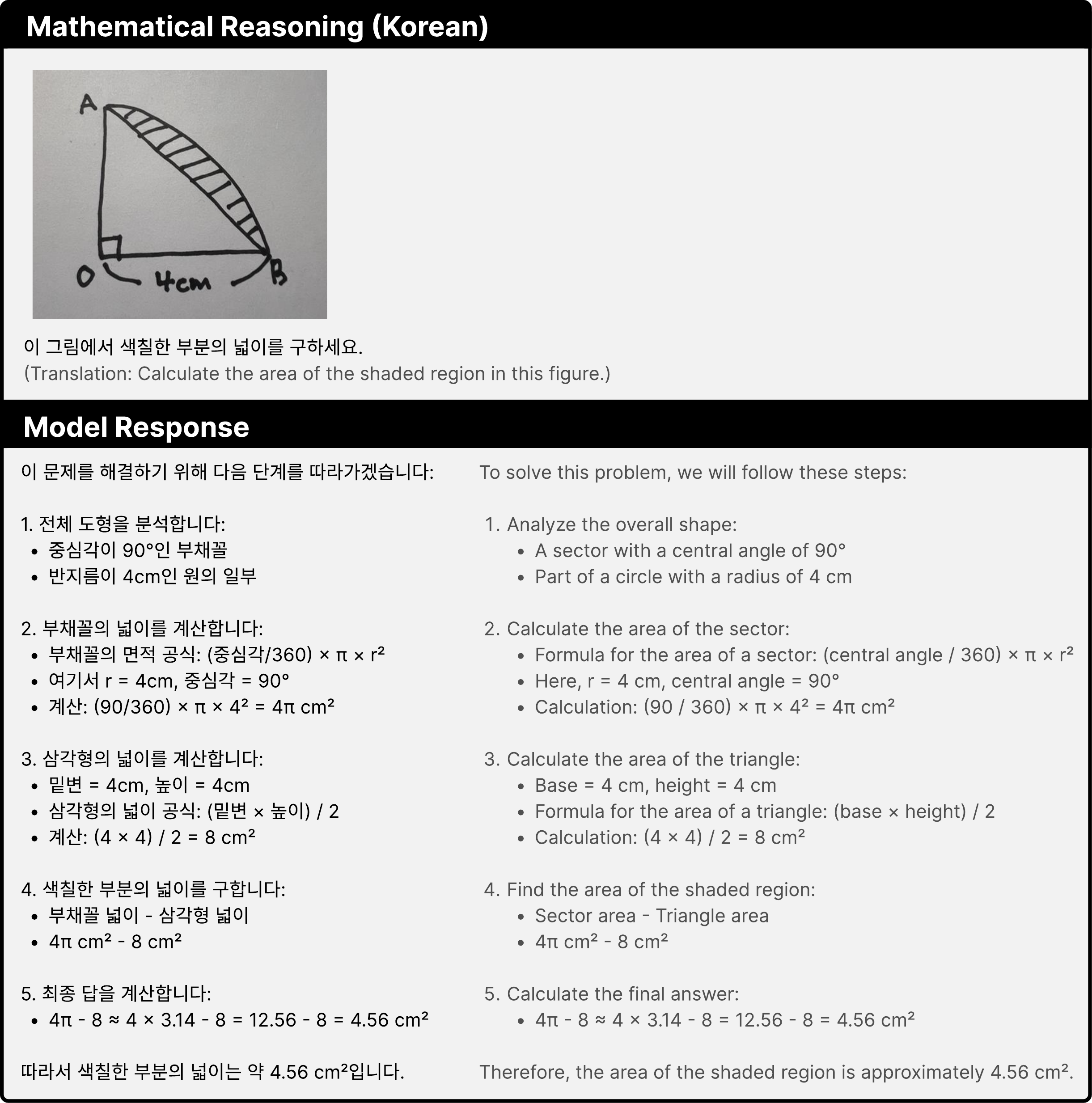} 
    \caption{Mathematical Reasoning Example (Korean)}
\end{figure}

\begin{figure}[H]
    \label{fig:calculation_ko}
    \centering
    \includegraphics[width=\textwidth]{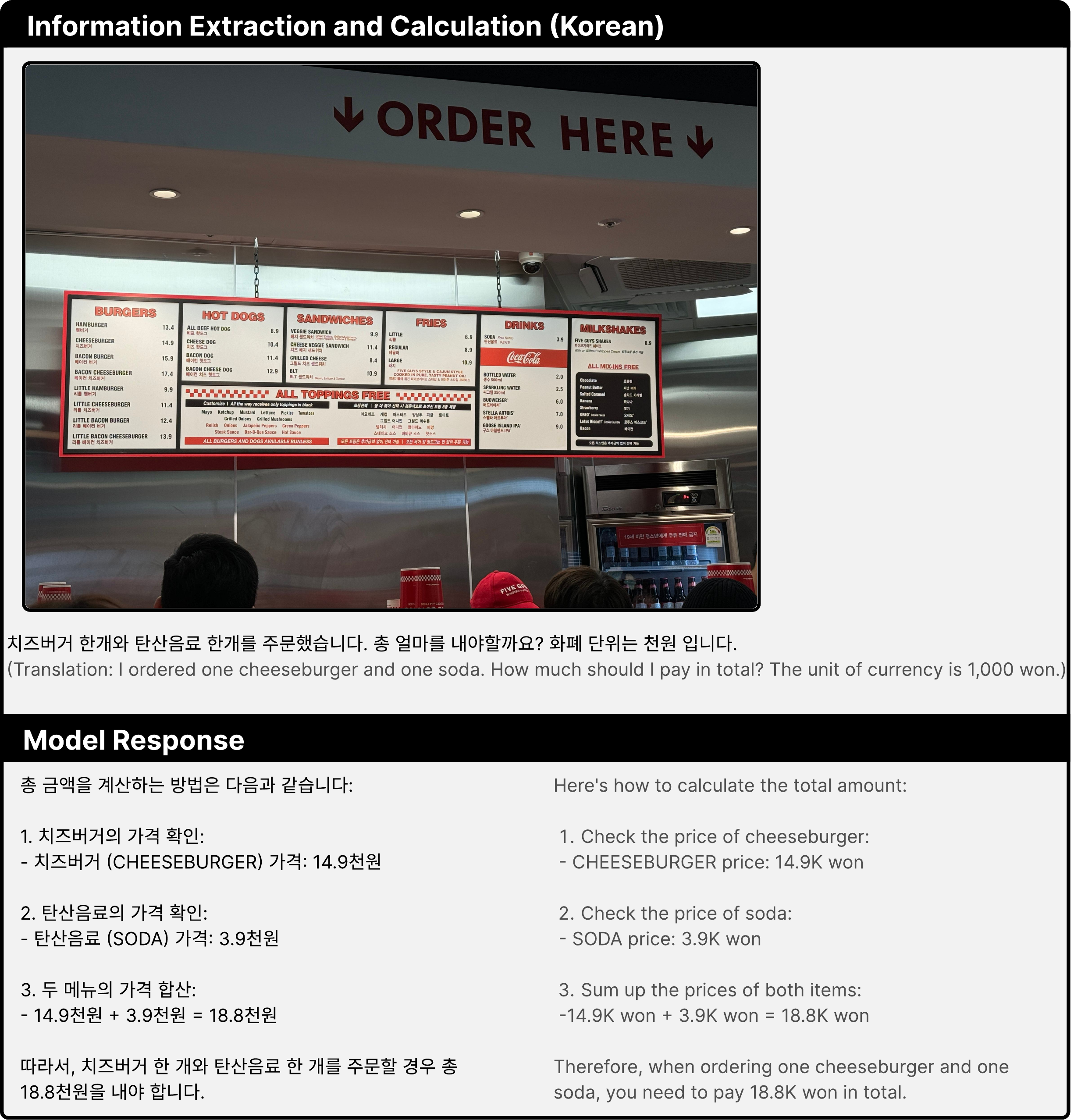} 
    \caption{Information Extraction and Calculation Example (Korean)}
\end{figure}

\begin{figure}[H]
    \label{fig:grounding_ko}
    \centering
    \includegraphics[width=\textwidth]{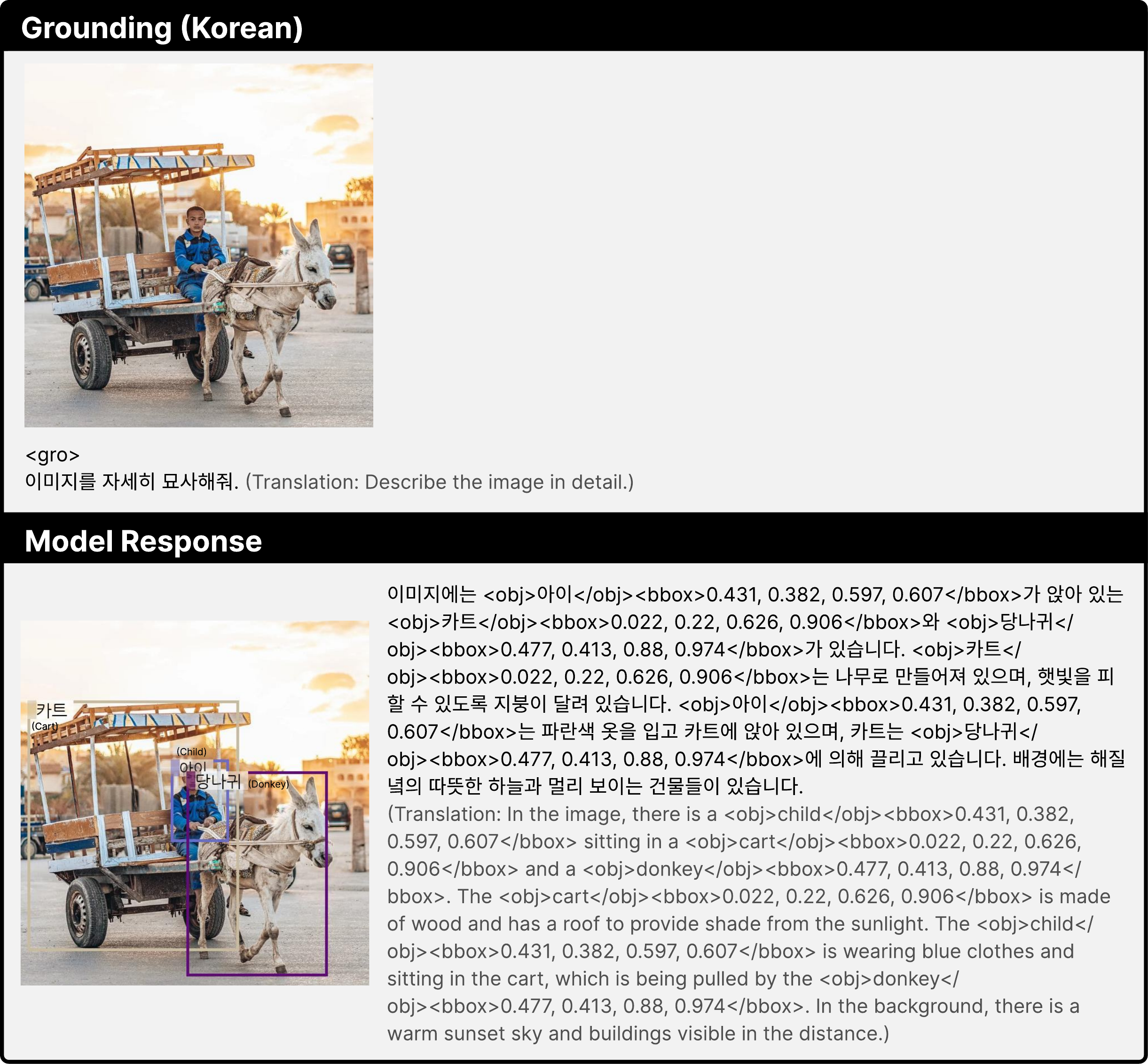} 
    \caption{Grounding Example (Korean)}
\end{figure}

\begin{figure}[H]
    \label{fig:grounding_en}
    \centering
    \includegraphics[width=\textwidth]{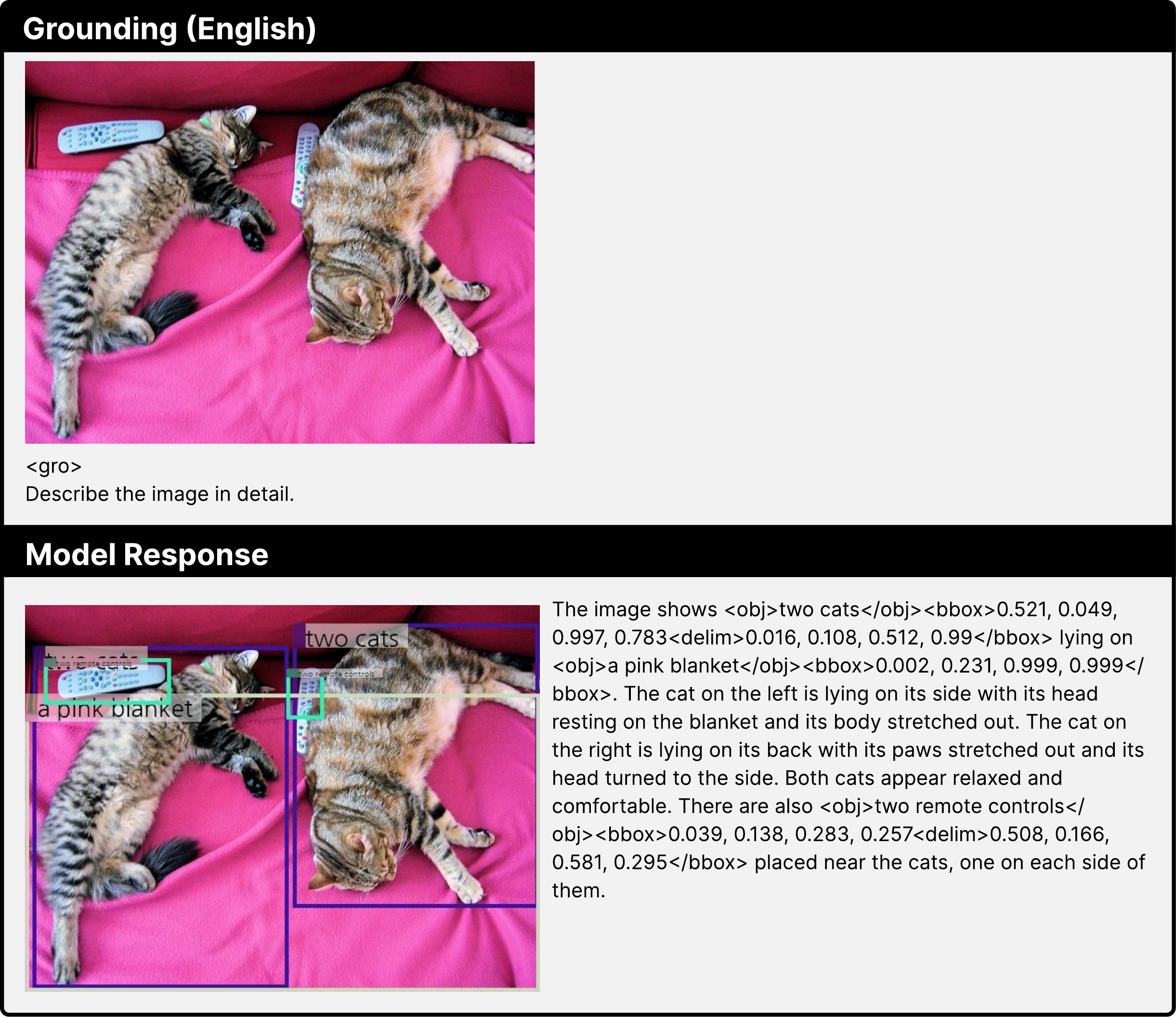} 
    \caption{Grounding Example (English)}
\end{figure}

\begin{figure}[H]
    \label{fig:referring_ko}
    \centering
    \includegraphics[width=\textwidth]{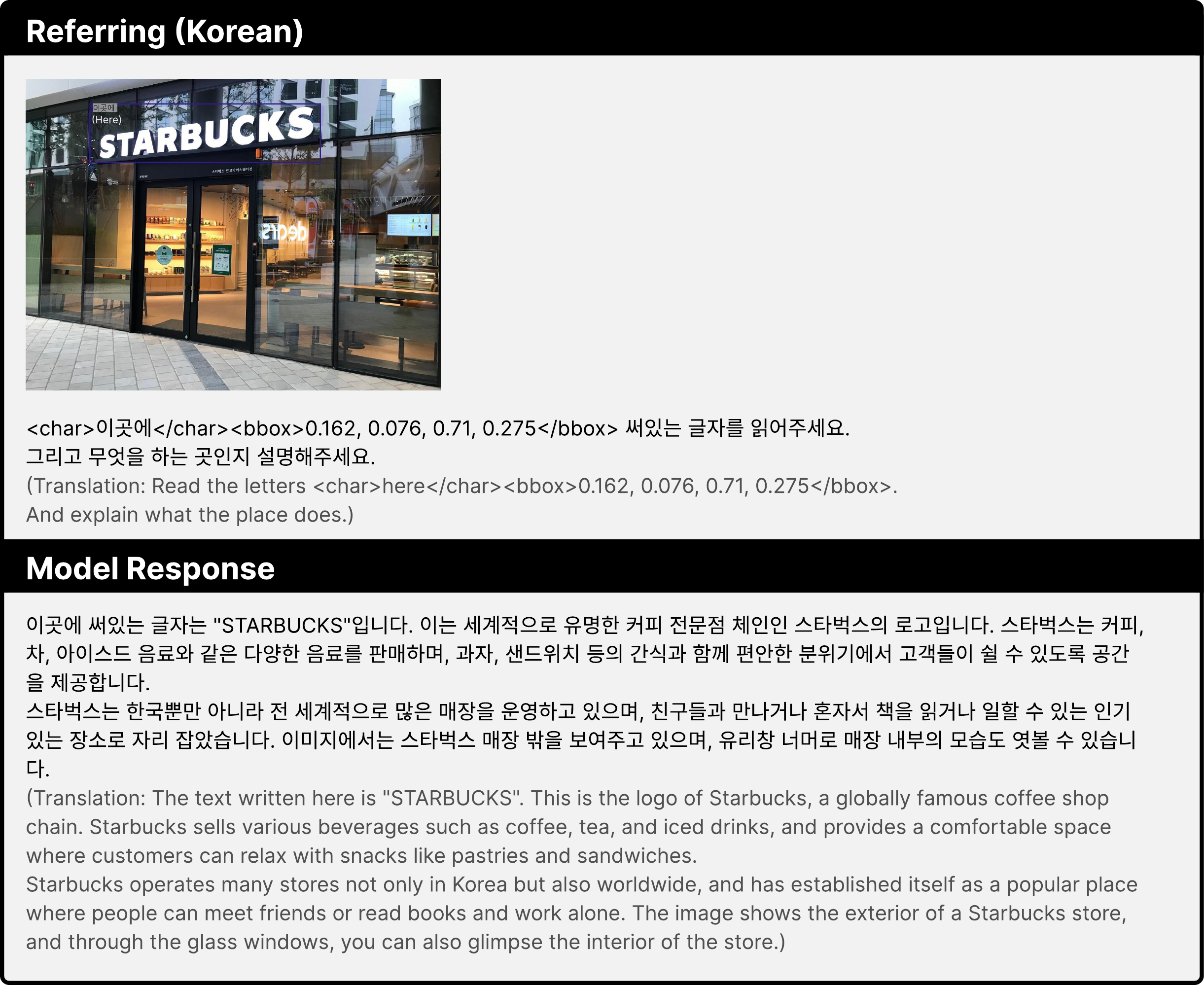} 
    \caption{Referring Example (Korean)}
\end{figure}

\begin{figure}[H]
    \label{fig:ocr_1}
    \centering
    \includegraphics[width=\textwidth]{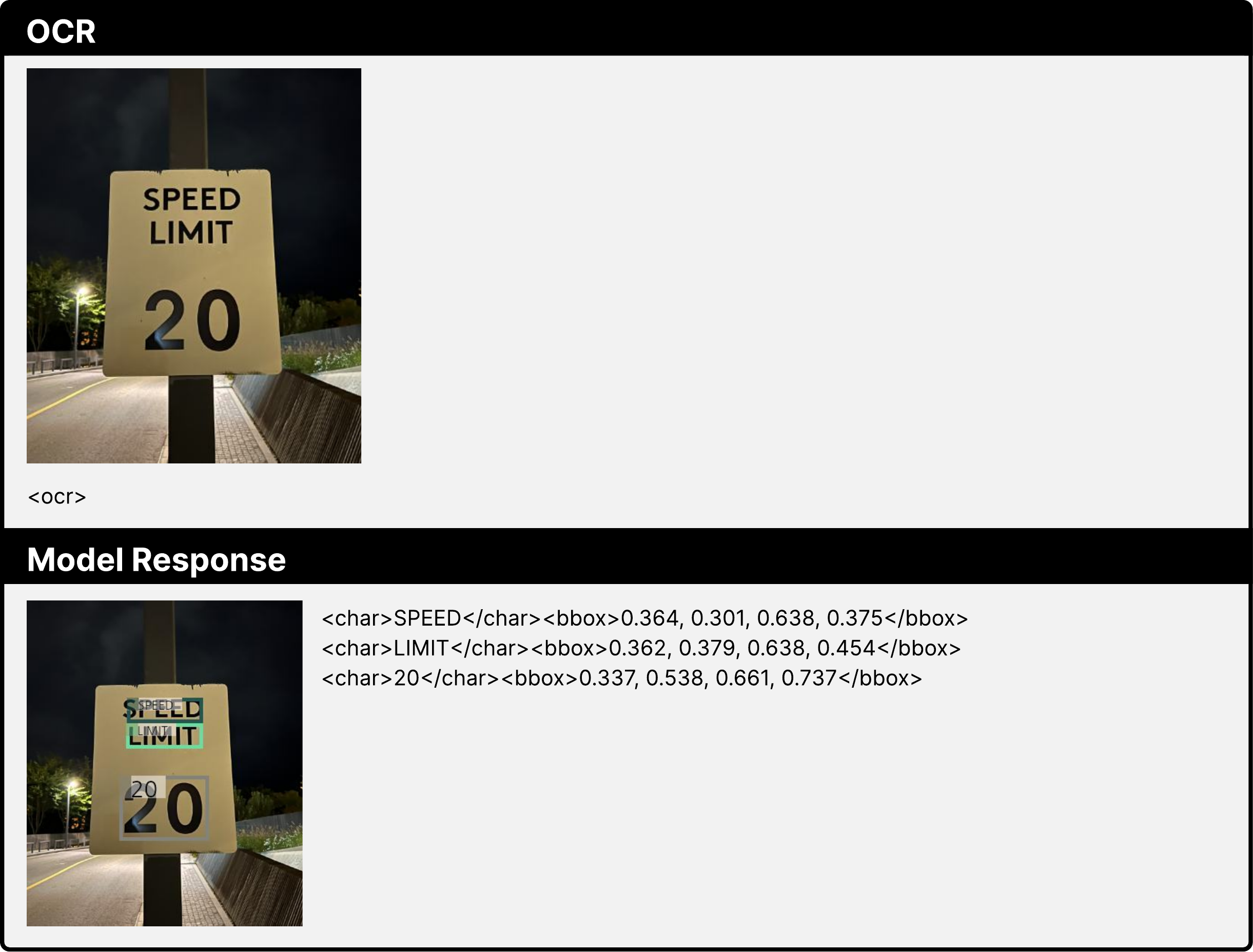} 
    \caption{OCR Example}
\end{figure}

\begin{figure}[H]
    \label{fig:ocr_2}
    \centering
    \includegraphics[width=\textwidth]{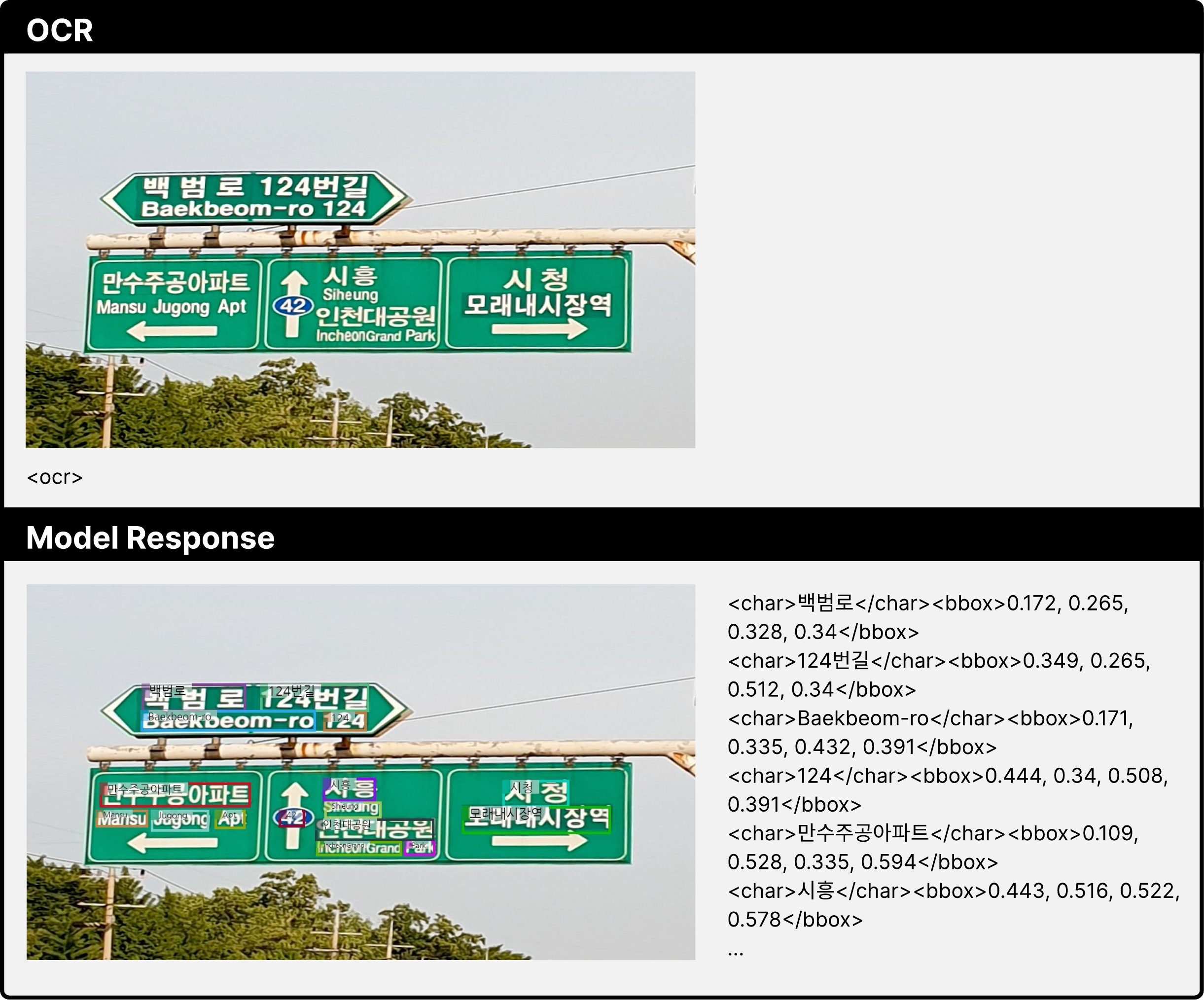} 
    \caption{OCR Example}
\end{figure}

\begin{figure}[H]
    \label{fig:summarization}
    \centering
    \includegraphics[width=\textwidth]{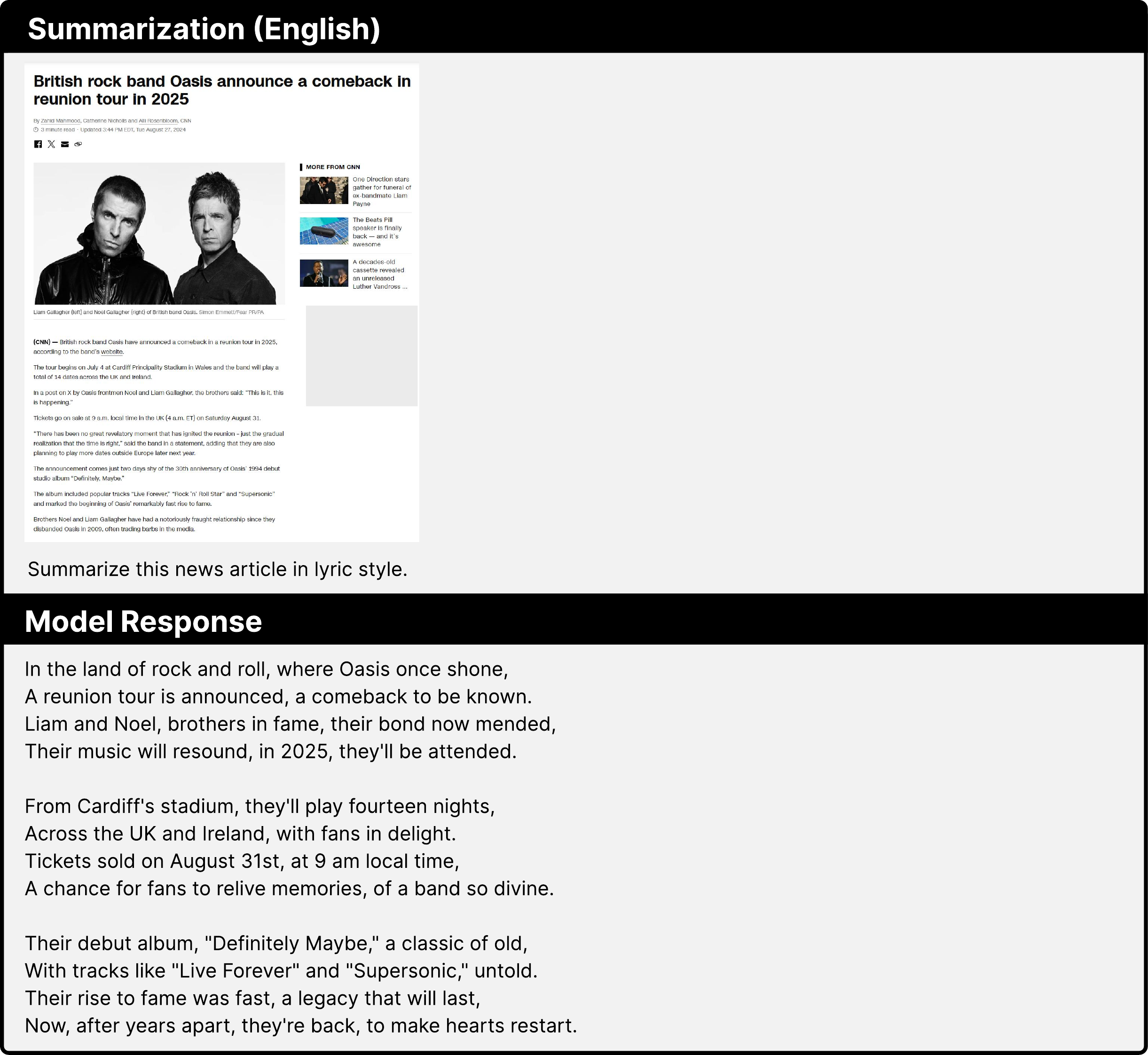} 
    \caption{Summarization Example (English)}
\end{figure}

\begin{figure}[H]
    \label{fig:recognition_ko}
    \includegraphics[width=0.99\textwidth]{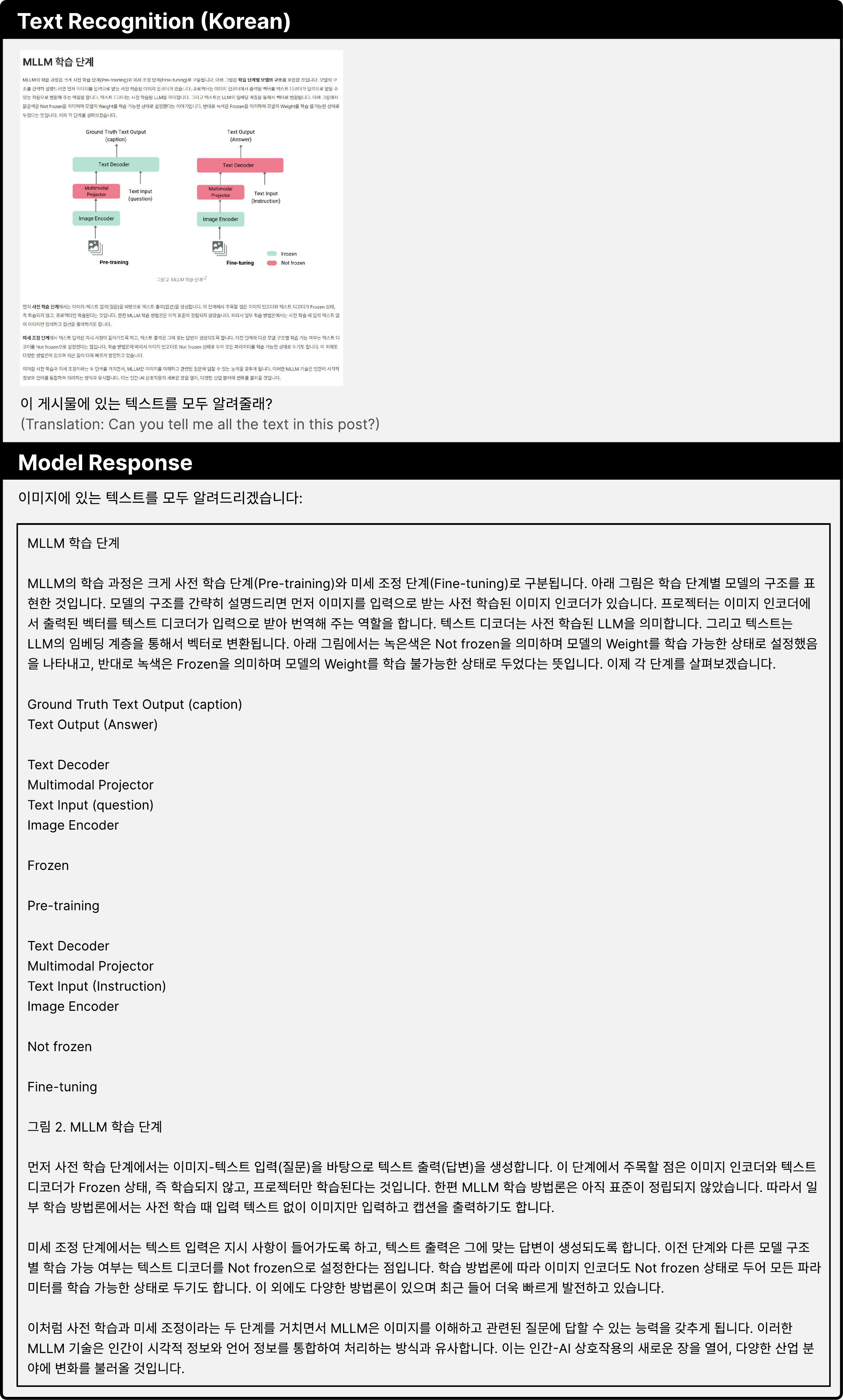} 
    \vspace{-2mm}
    \caption{Text Recognition Example (Korean)}
\end{figure}

\end{document}